%% file: main.tex
\documentclass[
  sora,
  numberedsections,
  onecolumn
]{mirrosarticle}

  \definecolor{ArenaInk}{HTML}{17151A}
  \definecolor{ArenaAccent}{HTML}{5F38FF}
  \definecolor{ArenaMuted}{HTML}{625E69}
  \definecolor{ArenaLine}{HTML}{D8DCE2}

  \newcommand{\AuthorEntry}[2]{%
  \mbox{\textbf{#1}}\hspace{1.05em plus 0.35em minus 0.35em}%
}

\input{authors.tex}

  \renewcommand{\authorlist}{%
    {\sffamily\color{ArenaInk}
     \fontsize{9.5}{13.2}\selectfont
     \leftskip=0pt \rightskip=0pt
     \parfillskip=0pt plus 1fil
      \emergencystretch=1.5em
     \AuthorList\par}%
  }

\renewcommand{\affiliationlist}{}

  \renewcommand{\contributionlist}{%
    \vspace{0.48cm}%
    {\color{ArenaLine}\hrule height 0.7pt}%
    \vspace{0.48cm}%
    \InstitutionLogos
  }

\usepackage[utf8]{inputenc}
\usepackage{amsmath}
\usepackage{amssymb}
\usepackage{amsfonts}
\usepackage{booktabs}
\usepackage{bm}
\usepackage{graphicx}
\usepackage{wrapfig}
\usepackage{xspace}
\usepackage{tikz}
\usetikzlibrary{arrows.meta,positioning}
\usepackage{capt-of}
\usepackage{url}
\usepackage{xcolor}
\usepackage{microtype}
\usepackage{enumitem}
\usepackage{hyperref}

\hypersetup{
  colorlinks=true,
  linkcolor=MirrosPurpleSoft,
  urlcolor=MirrosPurple,
  citecolor=MirrosPurpleSoft,
}

\setlist[itemize]{leftmargin=*}
\setlist[enumerate]{leftmargin=*}
\input{math_commands.tex}

\title{HarnessEval-W: Agentifying the Evaluation of \\ Visual Worlds}
\runningtitle{HarnessEval-W: Agentifying the Evaluation of Visual Worlds}
\date{\today}
\metadata[Blog]{\url{https://mirros.ai/blog/harnesseval}}
\metadata[Code]{\url{https://github.com/mirros-lab/harnesseval-w}}
\metadata[Project Page]{\url{https://mirros-lab.github.io/HarnessEval-W}}
\input{sections/0_abstract}

\begin{document}
\maketitle

\input{sections/1_introduction}
\input{sections/2_related}
\input{sections/3_method}

\input{sections/4_case_construction}
\input{sections/5_experiments}
\input{sections/6_future_work}

\input{sections/7_conclusion}

\printbibliography

\appendix
\input{sections/X_appendix}

\end{document}

%% file: authors.tex
\newcommand{\AuthorGroupLabel}[1]{%
  {\sffamily\bfseries\color{ArenaMuted}\fontsize{8}{9.8}\selectfont #1}\par
  \vspace{0.12cm}%
}
\newcommand{\CoFirstAuthorMark}{\textsuperscript{\textcolor{ArenaAccent}{$^*$}}}
\newcommand{\CoProjectLeaderMark}{\textsuperscript{\textcolor{ArenaAccent}{\ensuremath{\dag}}}}
\newcommand{\AuthorNotes}{%
  \par\vspace{0.08cm}%
  {\sffamily\color{ArenaMuted}\fontsize{7.4}{9}\selectfont
  \CoFirstAuthorMark Co-First Authors\hspace{1.2em}%
  \CoProjectLeaderMark Co-Project Leads}\par
}

\newcommand{\OrderedAuthorList}{%
  \AuthorEntry{Weiliang Chen\CoFirstAuthorMark}{}%
  \AuthorEntry{Haowen Sun\CoFirstAuthorMark}{}%
  \AuthorEntry{Jun Gao\CoFirstAuthorMark\CoProjectLeaderMark}{}%
  \AuthorEntry{Jiawei Chi}{}%
  \AuthorEntry{Hanyang Wang}{}%
  \AuthorEntry{Qiyu Dai}{}%
  \AuthorEntry{Yihao Li}{}%
  \AuthorEntry{Hao Li}{}%
  \AuthorEntry{Jingnan Gao}{}%
  \AuthorEntry{Yi-Hsin Hung}{}%
  \AuthorEntry{Xingzhuo Guo}{}%
  \AuthorEntry{Shangchen Miao}{}%
  \AuthorEntry{Zhiyuan Shi}{}%
  \AuthorEntry{Xiang Li}{}%
  \AuthorEntry{Fengrui Tian}{}%
  \AuthorEntry{Weihua Du}{}%
  \AuthorEntry{Ziqi Huang}{}%
  \AuthorEntry{Shenyuan Gao}{}%
  \AuthorEntry{Siqiao Huang}{}%
  \AuthorEntry{Mingyu Liu}{}%
  \AuthorEntry{Yifei Li}{}%
  \AuthorEntry{Shizun Wang}{}%
  \AuthorEntry{Xi Wang}{}%
  \AuthorEntry{Tianqi Zhang}{}%
  \AuthorEntry{Xue Luo}{}%
  \AuthorEntry{Yixin Ren}{}%
  \AuthorEntry{Jinshan Ren}{}%
  \AuthorEntry{Xiaoyang Shen}{}%
  \AuthorEntry{Xiaobo Hu}{}%
  \AuthorEntry{Zhiyang Dou}{}%
  \AuthorEntry{Mingyu Ding}{}%
  \AuthorEntry{Yichao Yan}{}%
  \AuthorEntry{Xinchao Wang}{}%
  \AuthorEntry{Yizhou Wang}{}%
  \AuthorEntry{Shilong Liu}{}%
  \AuthorEntry{Wenzhao Zheng}{}%
  \AuthorEntry{Yueqi Duan}{}%
  \AuthorEntry{Yuan Gong}{}%
  \AuthorEntry{Ziwei Liu}{}%
  \AuthorEntry{Ming-Yu Liu}{}%
  \AuthorEntry{Jialong Wu\CoProjectLeaderMark}{}%
  \AuthorEntry{Jiangran Lyu\CoProjectLeaderMark}{}%
  \AuthorEntry{Fangfu Liu\CoProjectLeaderMark}{}%
}

\newcommand{\AuthorList}{%
  \AuthorGroupLabel{Authors}%
  \OrderedAuthorList\par
  \AuthorNotes
}

\newcommand{\InstitutionLogos}{%
  \noindent\makebox[\textwidth][c]{%
    \includegraphics[width=\textwidth]{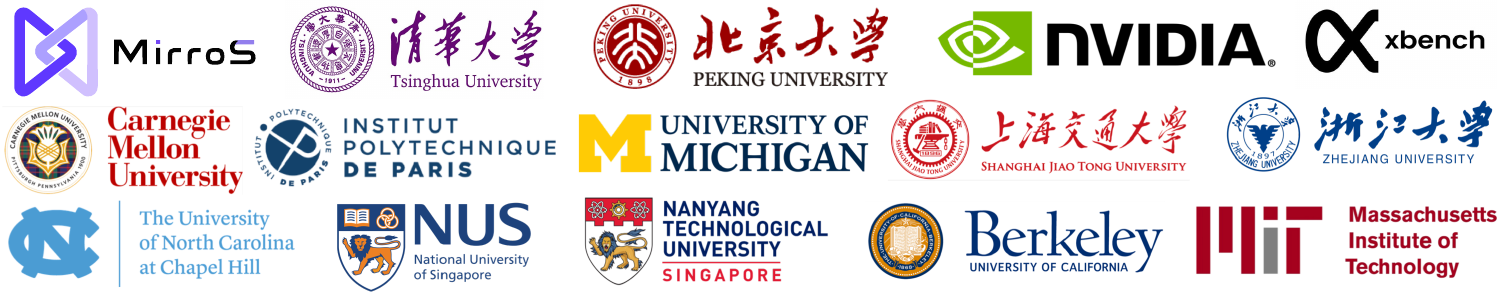}%
  }\par
}

%% file: math_commands.tex
\usepackage{amsmath,amsfonts,bm}

\def\eqref#1{equation~\ref{#1}}

\def\1{\bm{1}}

\newcommand{\ourmodel}{HarnessEval-W\xspace}

%% file: sections/0_abstract.tex
\begin{abstract}

A benchmark should deliver more than a scalar score: what makes an evaluation trustworthy is the reasoning that justifies the score. This is especially critical for world models, where judging a rollout requires understanding whether physics, causality, and world state evolve correctly. Humans spot such violations naturally, yet no existing benchmark automates this capability: metrics are computed brute-force, leaving no reasoning chain that can be examined or verified. We introduce \textbf{\ourmodel}, an agentified evaluation pipeline that brings the harness paradigm from the LLM ecosystem to world model benchmarking. Rather than applying a fixed rubric, \ourmodel interprets the context of each evaluation case, decomposes the evaluation question into measurable subproblems, and spawns specialized sub-agents, each equipped with tailored context and diagnostic tools to reason over its own subproblem. The parent agent then validates the gathered evidence and summarizes it into the final verdict. This hierarchical workflow turns every evaluation into a transparent evidence tree whose complete reasoning chain justifies the result. We apply \ourmodel to 18 representative world models over 330 evaluation cases. Its judgments closely align with human preferences while providing verifiable, fine-grained diagnoses of every generated rollout. We open-source the full pipeline as a live benchmark and invite the broad community to contribute to grow new skills and evaluation cases as world models evolve.

\end{abstract}

%% file: sections/1_introduction.tex
\section{Introduction}
\begin{figure}[t]
    \centering
    \includegraphics[width=\linewidth]{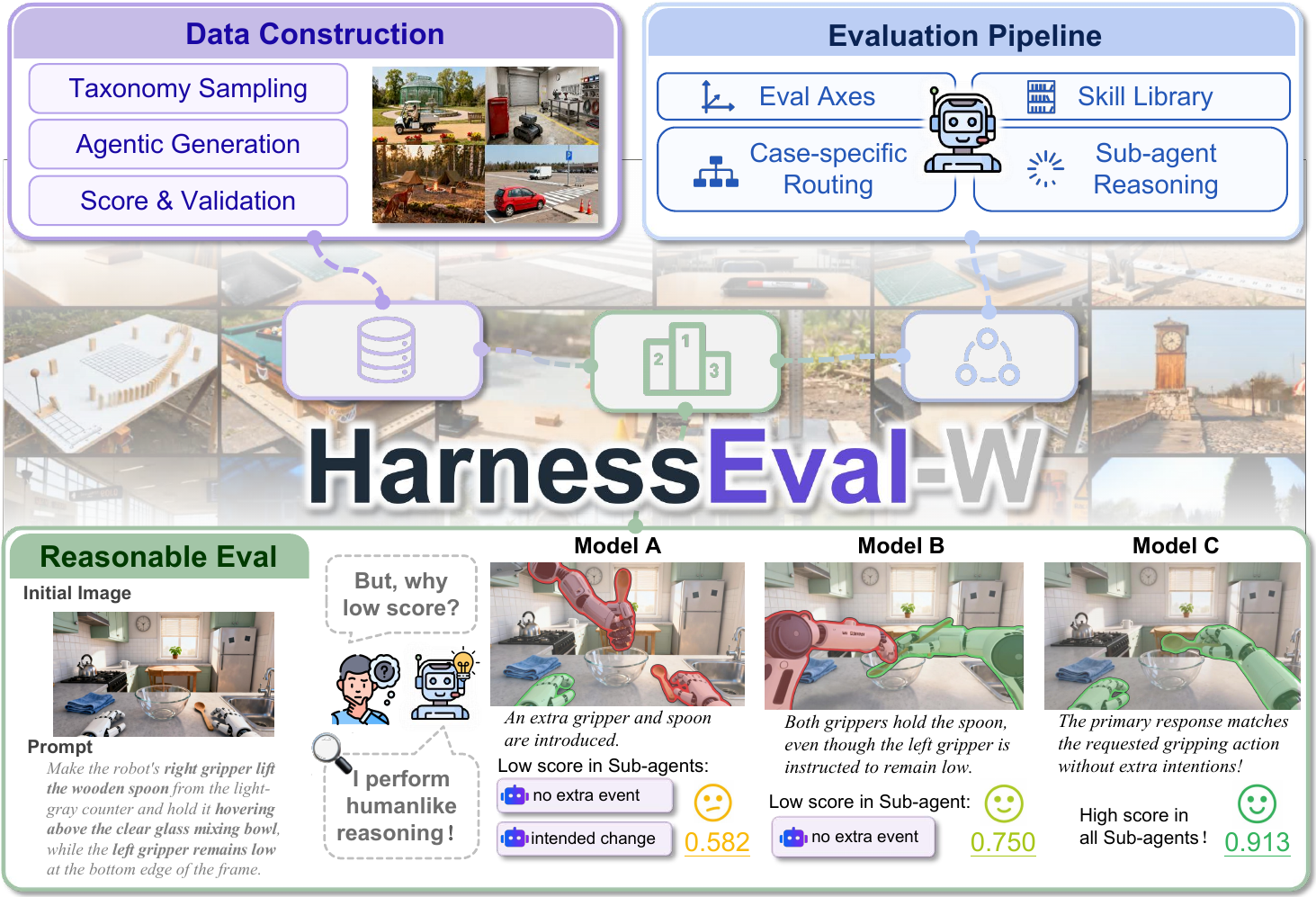}
    \caption{\textbf{\ourmodel}. The agentified benchmarks for interactive world models. Given an evaluation case, \ourmodel routes it to the appropriate skills, decomposes each skill into measurable sub-questions answered by specialized sub-agents, and aggregates the validated evidence into a final score that traces back to the exact sub-questions that failed.}
    \label{fig:teaser}
\end{figure}

Benchmarks set the North Star of technology development, guiding the frontier of exploration. Yet, evaluating world models remains a fragile practice, particularly when assessing physical causality, geometric consistency, and observation realism in the generated videos. While humans can easily analyze and spot the artifacts in the generated content, such a capability has never been successfully automated and achieved in existing benchmarks for world models. As a consequence, existing evaluation benchmarks offer limited persuasive power: the reported scores can be neither explained nor verified, and provide little insight into where and why a model fails.

In this paper, we bring a concept from the LLM ecosystem to the benchmark community: the \textbf{evaluation harness}. A harness is more than a code wrapper; it provides the robust scaffolding to power fully agentic pipelines by formalizing complex human workflows, such as evidence gathering, tool use, and reasoning. \textbf{Human evaluation is also a workflow}, and can therefore be harnessed. When humans evaluate a generated world, we do not merely glance at the static image; we intuitively locate the object, track object permanence across time, and verify causal relationships as well as geometric constraints.
We conceptualize this workflow as an agentic pipeline: a benchmark agent that actively interprets the intent of the actions performed in the world model, hunts for spatiotemporal evidence in the generation, and spawns specialized sub-agents, each equipped with dedicated context and diagnostic tools to investigate a different aspect of the model's output. This design moves beyond the static Q\&A probing of existing benchmarks and instead produces an inspectable reasoning sequence behind every score.

To realize this vision of agentic evaluation, we introduce \ourmodel: our agentic benchmark to evaluate interactive world models. Evaluating a world model is highly context-dependent: every test case instantiates a unique world with different physical actions, temporal structures, and observation states, so no fixed rubric fits all cases. \ourmodel therefore operates hierarchically. At the top level, it interprets the context of each evaluation case and routes it to the skills that can legitimately evaluate it. Once a skill is assigned, the evaluation question is further decomposed into measurable subquestions, each delegated to a suitable sub-agent or tool: when evaluating a collision, for instance, the agent tracks bounding boxes, verifies temporal intersection, and estimates velocity. The parent agent validates the evidence returned by all sub-agents before merging it into the final verdict, and this pipeline can run recursively as sub-agents spawn their own investigations. The outcome is not merely a scalar score but a transparent \emph{evidence tree}, which meticulously records exactly what was tested, which specific tool supplied the visual grounding, and the complete logical chain that justifies the final score.

In \ourmodel, we construct a benchmark of 330 evaluation cases organized around the three core functions of a world model: rendering observations from the current state (Observation Quality), updating the state under interventions (Transition Correctness), and maintaining the state over time (World Persistence). The cases are authored by an agentic construction pipeline that samples diverse initial worlds from a structured scene taxonomy, grounds concrete actions in the generated scenes, and validates every candidate case before acceptance. We evaluate 18 representative world models, spanning general-purpose video generators and interactive world simulators. The evaluation ordering from \ourmodel aligns well with human preferences. Our analysis further reveals that adapting a video generator into a world model redistributes capability across settings rather than uniformly improving it.

We believe the future of benchmarking is no longer a black-box metric with a static rubric, but an intelligent evaluator that decomposes the problem, assembles the right tools, and reasons over every evaluation case. We open-source \ourmodel as a living benchmark: an executable agentic system that grows new skills and evaluation cases as world models evolve. We invite the broader community to contribute to this agentic benchmark workflow together. 

%% file: sections/2_related.tex
\section{Related Works}

\subsection{Video Generation and Interactive World Models}

Modern video generation models have become increasingly capable of producing high-fidelity and temporally coherent sequences from text or image conditions. The field has progressed from U-Net-based approaches such as Video Diffusion Models~\citep{ho2022video} and latent diffusion systems such as Align Your Latents~\citep{blattmann2023align} to scalable diffusion transformer architectures, including CogVideoX~\citep{yang2024cogvideox}, HunyuanVideo~\citep{kong2024hunyuanvideo}, and Wan~\citep{wan2025wan}. These advances have improved visual fidelity, prompt adherence, temporal coherence, and generation horizons, providing strong visual priors for simulating observable worlds. Nevertheless, most video generators remain designed for open-loop synthesis, producing clips from fixed conditions without repeatedly incorporating external actions or maintaining an evolving world state. This limitation separates the synthesis of plausible visual sequences from the modeling of worlds that respond coherently to interaction.

Interactive video world models bridge this gap by conditioning future observations on actions and placing generation within a closed interaction loop. Classical approaches~\citep{ha2018worldmodels,hafner2019dreamer, hafner2023dreamerv3} learn compact latent dynamics for planning and control, whereas recent generative systems predict action-conditioned futures directly in pixel space. This direction includes game-centered systems such as Genie~\citep{bruce2024genie}, GameNGen~\citep{valevski2024gamengen}, DIAMOND~\citep{alonso2024diamond}, and MineWorld~\citep{guo2025mineworld}, as well as more open-ended models supporting language, camera, and keyboard-and-mouse control, including YUME~1.5~\citep{mao2025yume15}, HY-World~1.5 (WorldPlay)~\citep{sun2025worldplay}, LingBot-World~\citep{gao2026lingbotworld}, and Matrix-Game~3.0~\citep{wang2026matrixgame3}. Despite their different control interfaces, interactive world models must be assessed not only by the reliability of their generated observations, but also by whether actions produce the intended semantic, geometric, and physical transitions and whether visible and offscreen state remains coherent over long rollouts. Accordingly, we structure {HarnessEval-W}'s evaluation around three complementary axes: {Observation Quality}, {Transition Correctness}, and {World Persistence}.

\subsection{Benchmarking Video Generators and Interactive Worlds}
Benchmarking an interactive world begins with {visual and temporal reliability}, because later judgments of dynamics and persistence depend on usable rendered observations. VBench \citep{huang2024vbench} decomposes video quality across imaging, aesthetics, temporal stability, subject and background consistency, motion, and prompt consistency, with dimension-wise human validation. EvalCrafter \citep{liu2024evalcrafter} complements this taxonomy with visual, content, motion, and text-video metrics calibrated by user judgments. FETV \citep{liu2023fetv} uses temporally aware prompts and human ratings to expose metric weaknesses, while Video-Bench and VBench++ \citep{han2025videobench,zheng2025vbenchplusplus} extend coverage through MLLM judging, image-to-video generation, and trustworthiness.

Reliable rendering is only the first step, as world realism also requires {physically and behaviorally plausible content}. VBench-2.0 \citep{zheng2025vbench2} covers human fidelity, controllability, creativity, physics, and commonsense, while VideoPhy~\citep{bansal2024videophy}, PhyGenBench~\citep{meng2025phygenbench}, and PhyWorldBench~\citep{gu2026phyworldbench} test material interactions, physical-law reasoning, and composite or anti-physical scenarios. Together, they distinguish realistic appearance from behavior compatible with the depicted world.

Interactive-world benchmarks then make actions explicit and assess {whether controls produce the intended changes}. WorldScore \citep{duan2025worldscore} compares 3D, 4D, and video generation on camera-specified next scenes, while related benchmarks test instruction and physics adherence and whether futures support intended controls \citep{li2025worldmodelbench,qin2025worldsimbench}. WorldMark and complementary suites standardize interfaces and evaluate generation, trajectories, memory, and causal effects \citep{xu2026worldmark,fang2026iworldbench,wu2026omniworldbench}. Downstream benchmarks such as WorldArena cover data generation, policy evaluation, planning, closed-loop success, and multi-turn interaction under human-validated quality, adherence, consistency, and physics metrics \citep{shang2026worldarena,zhang2026worldinworld,ying2026wbench}.

Evaluation next asks whether locally valid transitions preserve {one consistent, evolving world} over longer horizons. WorldScore and WorldMark evaluate geometric, appearance, and state consistency through camera specifications, reprojection, and long trajectories \citep{duan2025worldscore,xu2026worldmark}. Revisitation suites expose drift through return paths and long-context reconstruction \citep{fang2026iworldbench,ying2026wbench,ye2026mind}, but cannot determine {whether hidden state evolves offscreen}. MemoBench and related benchmarks test offscreen updates, re-observation correctness, and extended action, physics, and memory rollouts \citep{chen2026memobench,lu2026wrbench,xu2026worldroambench}.

Across all evaluation targets, {evaluation reliability} is a cross-cutting requirement. MUSIQ and related tools extract frame-quality, camera-motion, return-observation, and mask evidence \citep{ke2021musiq,li2025megasam,fu2023dreamsim,ravi2025sam2}, while VLMEvalKit and other toolkits support reproducible pipelines \citep{duan2024vlmevalkit,liang2023helm,bandel2024unitxt}. Because such pipelines are fixed, interactive worlds require adaptive applicability, execution, and aggregation, motivating agentic evaluation.

\begin{figure}[t]
    \centering
    \includegraphics[width=\linewidth]{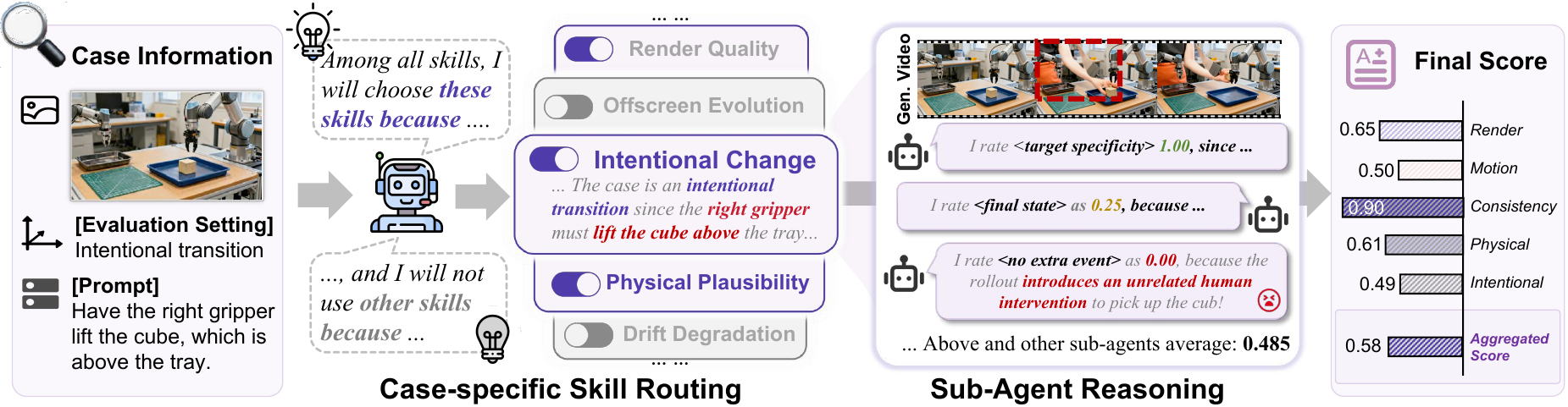}
    \caption{\textbf{Overview of the \ourmodel evaluation pipeline.} Given a case with its prompt and evaluation setting, the agentic planner routes the case to applicable skills from the skill library, recording an evidence-grounded reason for every activated and skipped skill (\textit{e.g.}, the Offscreen Evolution Verifier is skipped because all requested actions remain visible). Each activated skill spawns sub-agents that inspect the world model rollout for specific sub-questions, such as target visibility and final state validity; here, the Intentional Change Verifier detects an unrelated human intervention picking up the cube and zeroes the no-extra-event score. The collected evidence is finally aggregated into per-dimension scores and an interpretable final score.}
    \label{fig:pipeline}
\end{figure}

\subsection{Agentic Evaluation}

Most existing benchmarks predefine what to test, how to extract evidence, and how to validate and aggregate scores. In interactive worlds, these choices are state dependent because available controls and earlier outcomes determine what can be tested next. This motivates {{agentic evaluation}}, in which the evaluator selects, executes, and revises probes as evidence accumulates. In LLM evaluation, model-based judges~\citep{zheng2023judging} provide a useful precedent by applying natural-language criteria, although they typically assess supplied outputs under fixed prompts. More recent agentic evaluators extend this role to test construction, evidence acquisition, tool use, memory, and verification \citep{you2026agentasjudge,zhuge2025agentasjudge,gou2025mind2web2,brown2025taskelicitation,shen2026oneeval,wang2026agenticeval}. In this setting, agency lies in controlling the evaluation trajectory rather than merely replacing a metric with a model-based judge.

In visual generation, this procedural view extends to selecting which prompts to sample, which regions or frames to inspect, and which tools to apply. Adaptive testing methods \citep{zhang2025evaluationagent,tian2026openevaluationagent,song2026vqqa} support this process by revising prompts and subsequent probes using feedback from sampled outputs. Tool-grounded image evaluators \citep{qin2024evaluating,wang2025cigeval,chen2026edival} organize fine-grained evidence across generation and multi-turn editing, while temporally aware video evaluators~\citep{yang2025videogen,zeng2026videoargus} combine structured criteria with localized evidence. Reusable evaluation frameworks~\citep{zhang2026rewardharness} further externalize evaluation knowledge as revisable libraries of tools and skills. Together, these methods make prompts, rubrics, and evidence acquisition adaptive, but largely remain within prompt-conditioned image or video generation and editing.

HarnessEval-W extends this adaptive logic to interactive worlds through reusable evaluation skills. Each skill defines its applicability, evidence, scoring, validation, and aggregation, allowing probes to be selected and revised as a rollout unfolds. This separation keeps unsupported or invalid measurements out of the score while preserving genuine model failures as negative evidence. Such safeguards are important because insufficient evidence, unreliable state verification, and benchmark design choices can distort evaluation outcomes \citep{shi2026ajbench,zhu2025rigorousagentbenchmarks}. HarnessEval-W thus unifies Observation Quality, Transition Correctness, and World Persistence within an auditable, adaptive process.

%% file: sections/3_method.tex
\section{HarnessEval-W: Agent Evaluation Harness}

\subsection{Overview}
Figure~\ref{fig:pipeline} illustrates the overall pipeline of our agentic evaluation harness for interactive world models. The core idea of \ourmodel is to decompose the monolithic task of evaluation into distinct, actionable subproblems. We first formalize the definition of world models and decompose their capabilities into three evaluation axes: observation quality, transition correctness, and world persistence. Our hierarchical agentic workflow then evaluates world models along detail evaluation settings from these three axes.

\subsection{The Formulation of Interactive World Models}
Following common practice in the literature~\cite{hafner2023mastering,hafner2019dreamer,ha2018worldmodels}, we formulate an interactive world model as predicting future observations conditioned on historical observations and user-specified actions. Any model that supports this interface can be evaluated by our benchmark, including but not limited to bidirectional video diffusion models and autoregressive video models. Formally, given initial observations $\{o_i\}_{i=-T}^{0}$ and future actions $\{a_i\}_{i=0}^{t-1}$, we decompose the distribution of future observations $\{o_i\}_{i=1}^{t}$ as follows, leveraging the hidden states $\{s_i\}_{i=0}^{t}$ of the world model:
\begin{equation}
\boxed{
   P(o_1, \cdots, o_t \mid o_{-T}, \cdots, o_0;\, a_0, \cdots, a_{t-1})
   \propto P(s_0 \mid o_{-T}, \cdots, o_0) \prod_{i=1}^{t} S(o_i \mid s_i)\, T(s_i \mid s_{i-1}, a_{i-1}),
}
\label{eq:world_model_factorization}
\end{equation}
where the distribution of future observations is obtained by marginalizing the hidden states. Here $S(o_i \mid s_i)$ models the observation likelihood conditioned on the current state, $T(s_i \mid s_{i-1}, a_{i-1})$ refers to the state transition conditioned on the given action, and $P(s_0 \mid o_{-T}, \cdots, o_0)$ is the initial state distribution inferred from historical observations. An action can be an exploratory viewpoint change, an intentional change to a specified entity or event, or a physical intervention.

\subsection{What to Test: Evaluation Axes}
The factorization in Equation~(\ref{eq:world_model_factorization}) exposes three basic abilities of a world model: rendering observations from the current state ($S$), updating the state under actions ($T$), and maintaining a coherent state sequence over time ($\{s_i\}_{i=0}^{t}$). As illustrated in Table~\ref{tab:evaluation_axes}, we construct three corresponding evaluation axes with eight detail evaluation settings in \ourmodel.

\textbf{Observation Quality} refers to whether the rendered observation of the current world state is visually reliable. It covers perceptual quality, temporal coherence, and the readability of state-relevant entities, attributes, relations, and events. Observation quality reflects the action-agnostic generation capability of a world model and is the foundation on which all further evaluation rests.

\textbf{Transition Correctness} focuses on whether the state transition faithfully executes the requested actions at the appropriate time. We evaluate three transition types: \emph{Exploratory} transitions change the observer's viewpoint or position in the world; \emph{Intentional} transitions change a specified entity, relation, or event; and \emph{Physical} transitions test whether a physical intervention produces the corresponding dynamical response.

\textbf{World Persistence} evaluates whether the sequence of predicted states remains coherent as the world evolves. It includes three representative settings: \emph{Drift Resistance} tests whether overall layout, style, and appearance remain consistent across a long rollout; \emph{Revisit Consistency} tests whether a location or object remains the same after the observer leaves and returns; and \emph{Offscreen Evolution} examines whether an endogenous process continues while temporarily unseen. Persistence does not require the world to remain frozen. Instead, stable properties must stay invariant, while dynamic properties should continue to evolve consistently with actions and time.

\input{tables/tab_evaluation_axes.tex}

\subsection{How to Test: Hierarchical Agentic Evaluation}

The evaluation of world models is highly context-dependent: each test case is a distinct world with its own actions, temporal structure, and observable state, so a fixed rubric applied uniformly across cases would either ask irrelevant questions or miss case-critical context. We therefore design a {hierarchical} agentic workflow that generates a reasoning trace grounded in the context of each case, as shown in Figure~\ref{fig:pipeline}. Specifically, \ourmodel first routes each case to the appropriate skills according to the case context and evaluation intention. Each skill is an intelligent agent that decomposes its evaluation problem into measurable sub-questions, delegates them to sub-agents that interpret the world from specialized perspectives, and finally the parent agent validates and aggregates the collected evidence into the final score.

\begin{figure}[t]
    \centering
    \includegraphics[width=\linewidth]{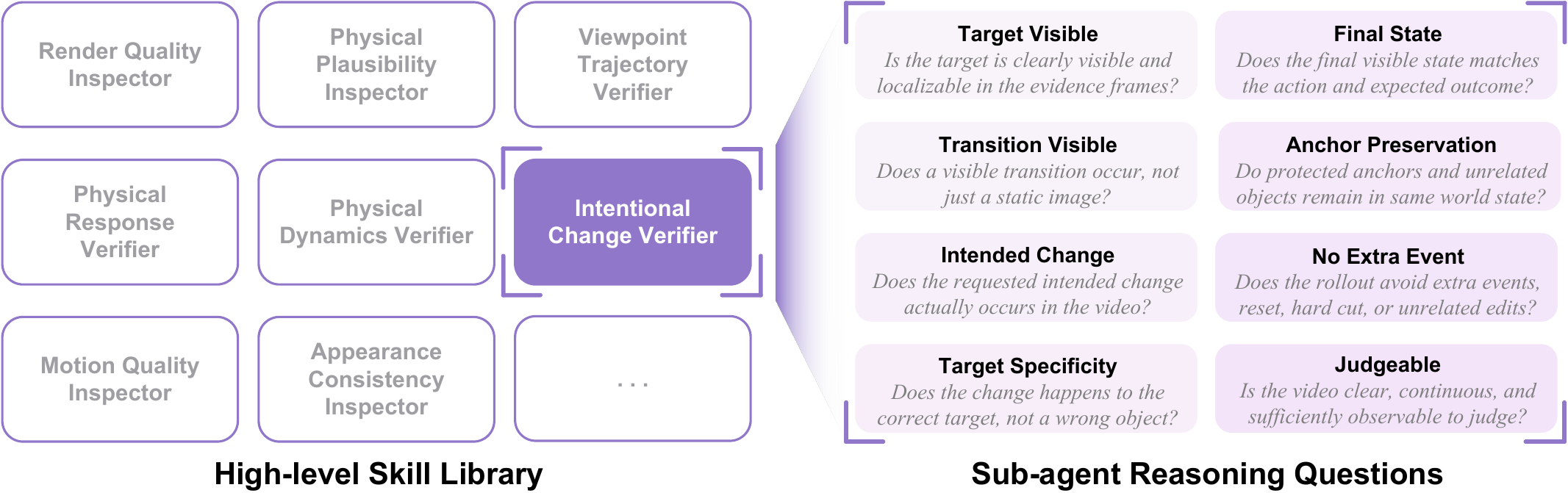}
    \caption{\textbf{Hierarchical structure of sub-agent reasoning for high-level skills.} The Intentional Change Verifier skill decomposes its evaluation into eight measurable sub-questions, each answered by a dedicated sub-agent that inspects the rollout evidence and returns a discrete score with its reasoning.}
    \label{fig:subagent_example}
\end{figure}

\textbf{Case-specific Skill Routing.} The evaluation begins by interpreting the case context, including the initial image, action prompt, and evaluation setting. Based on this context, \ourmodel routes the case to one or more reusable skills drawn from a predefined skill library as shown in Figure~\ref{fig:subagent_example} (left). 
Each high-level skill is tailored to a specific evaluation type, ensuring that we ask the right high-level questions for each unique world instead of a one-size-fits-all question. The skill library is designed to be extensible: new skills can be added as new evaluation needs arise, without changing the overall framework. Note that the routing process depends only on the case context and evaluation intention and is independent of the world model being evaluated, so the same routing is applied to all models on the same case, ensuring the evaluation is fair and unbiased.

\textbf{Sub-agent Reasoning.} Once a high-level skill is assigned, \ourmodel performs a granular evaluation via sub-agents rather than issuing a single holistic evaluation. Specifically, for each high-level skill, we further decompose the evaluation into a set of sub-questions that can be answered by specialized sub-agents.
Each sub-agent is then only responsible for one specific aspect of the evaluation, such as checking target visibility or confirming that a visible transition occurs, and returns a discrete score together with diagnostics. The parent skill agent then aggregates these outputs into a structured evidence tree and produces the case score. This hierarchical design ensures that the reasoning trace records not only the final answer but also the supporting evidence behind it. 

As shown in Figure~\ref{fig:subagent_example}, we take Intentional Change Verifier as an example. \ourmodel further decomposes this high-level skill through multiple sub-agents with detail questioning. Before querying these specific sub-agents, an extra sub-agent first predicts and reasons about the expected outcome from the case context. Guided by this specification, parallel sub-agents then inspect the rollout for target visibility, visible transition occurrence, and target correctness, while additional branches examine the final state, anchor preservation, extra events and overall judgeability. Since each sub-agent scores the rollout against its specific expectation, the final score remains interpretable: a low score can be traced back to a concrete failure, \textit{e.g.}, the wrong target or including extra unrelated events. We provide specific cases in Section~\ref{sec:experiments}.

The ultimate goal of our agentic workflow is not just to provide a scalar score indicating model performance, but a transparent evidence tree, which meticulously records exactly what was tested, which specific tool supplied the visual grounding, and the complete logical chain that justifies the final score. This interpretable paradigm further delivers actionable insights, helping researchers pinpoint exact failure modes and guiding future model development.

%% file: tables/tab_evaluation_axes.tex
\begin{table}[t]
\centering
\small
\setlength{\tabcolsep}{5pt}
\renewcommand{\arraystretch}{1.08}
\caption{\textbf{Evaluation axes of \ourmodel.} Derived from the factorization in Equation~(\ref{eq:world_model_factorization}), the three evaluation axes decompose into eight detailed settings, each anchored by the core world-state question it answers.}
\begin{tabular}{p{0.23\linewidth}p{0.27\linewidth}p{0.42\linewidth}}
\toprule
\textbf{Evaluation Axis} & \textbf{Detail Evaluation Settings} & \textbf{Core World-state Question} \\
\midrule
\multirow[c]{4}{=}{Observation Quality} & \multirow[c]{2}{=}{Render Quality} &
Is the rollout video coherent, stable, and sufficiently readable to serve as evidence? \\
& \multirow[c]{2}{=}{Physical Observation} &
Is each rendered frame from hidden state structurally and physically plausible? \\
\midrule
\multirow[c]{6}{=}{Transition Correctness} & \multirow[c]{2}{=}{Exploratory Transition} &
Does the requested viewpoint change occur while the world remains compatible? \\
& \multirow[c]{2}{=}{Intentional Transition} &
Does the specified target change while protected state remains stable? \\
& \multirow[c]{2}{=}{Physical Transition} &
Does a physical intervention produce the corresponding dynamical response? \\
\midrule
\multirow[c]{5}{=}{World Persistence} & Drift Resistance &
Do invariant objects survive a long rollout? \\
& \multirow[c]{2}{=}{Revisit Consistency} &
Does a place or entity remain compatible after leaving and returning? \\
& \multirow[c]{2}{=}{Offscreen Evolution} &
Does an unobserved endogenous process continue rather than freeze or reset? \\
\bottomrule
\end{tabular}

\label{tab:evaluation_axes}
\end{table}

%% file: sections/4_case_construction.tex
\section{HarnessEval-W Case Construction}
\label{sec:case_construction}

\begin{figure}[t]
    \centering
    \includegraphics[width=\linewidth]{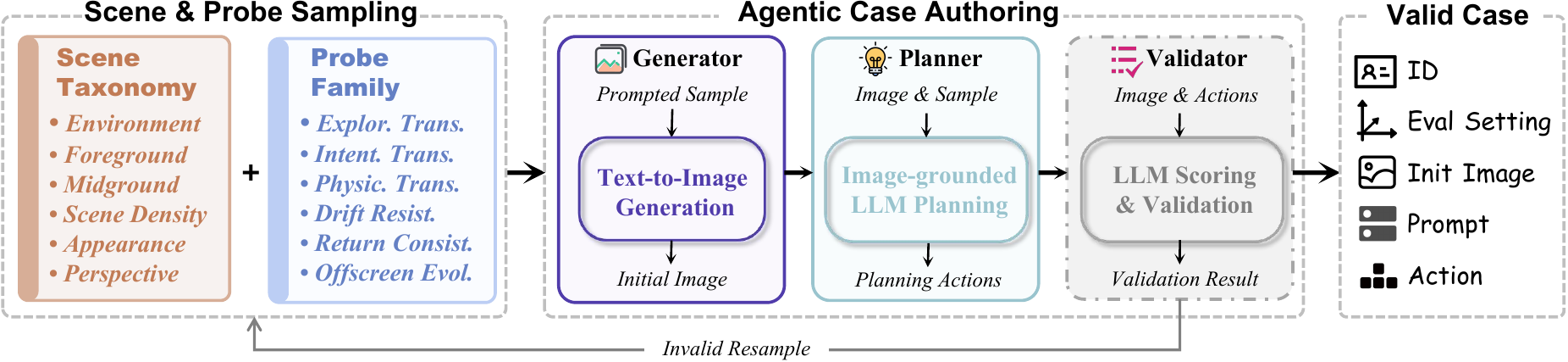}
    \caption{\textbf{Overview of the \ourmodel data construction pipeline.} We first sample an initial world setup from a predefined scene taxonomy and probe family, and then employ a series of agents for world generation, action planning, and case validation.}
    \label{fig:case_construction}
\end{figure}

\textbf{Overall Pipeline.} To probe the limits of world models, a benchmark must reflect the complexity and diversity of real-world environments, demanding a wide spectrum of evaluation cases. However, manually authoring such cases is hard to scale while keeping every case diagnosable and every intervention verifiable. In practice, we therefore need both real-world data and synthetic cases that are diverse and realistic. To this end, we design our agentic \ourmodel case construction pipeline, illustrated in Figure~\ref{fig:case_construction}. The pipeline first samples an initial world setup from a predefined scene taxonomy and probe family, and then employs a series of agents for world generation, action planning, and case validation.

\textbf{Scene Taxonomy Sampling.} To provide complete and diverse descriptions for world initialization, we develop a structured scene taxonomy with six complementary axes: 1) \emph{Environment} represents the basic background setting, organized into indoor (\textit{e.g.}, living room, office), outdoor (\textit{e.g.}, traffic intersection), and transitional (\textit{e.g.}, doorway, building entrance) categories; 2) \emph{Foreground} specifies the dominant visible entities that can be manipulated or serve as landmarks, such as people, vehicles, articulated objects, and physical apparatuses; 3) \emph{Midground} refers to the spatial structure in which an interaction takes place, such as a corridor, sidewalk, workbench, road, or shoreline; 4) \emph{Scene Density} controls the amount of clutter and objects while preserving the readability of case-critical entities; 5) \emph{Appearance} ranges from photorealistic and cinematic rendering to game-engine, stylized-3D, and anime domains; 6) \emph{Perspective} specifies the observer's relationship to the world, including first-person and third-person views.
During construction, a sampler selects a compatible combination of taxonomy values to instantiate a concrete world. It also checks basic semantic compatibility, rejecting combinations whose entities or spatial structures cannot support the selected interaction.

\textbf{Probe Family Assignment.} In addition to the scene taxonomy, we assign each case a probe family that specifies its interaction type and the evidence expected for evaluation. The six case families correspond to the Transition Correctness and World Persistence settings in Table~\ref{tab:evaluation_axes}, while Observation Quality is evaluated in every case and therefore does not constitute a separate family. Each probe family imposes its own requirements on the sampled scene: 1) \textit{Exploratory Transition} requires a traversable route, an open region, or stable spatial anchors that support a viewpoint or navigation change; 2) \textit{Intentional Transition} requires a clearly visible target with an identifiable initial state and a localized edit, relation change, or event; 3) \textit{Physical Transition} requires visible support, contact, and motion relationships, together with state variables whose response can be inspected; 4) \textit{Drift Resistance} requires the overall layout, style, and appearance remain consistent across a long rollout; 5) \textit{Revisit Consistency} requires a recognizable location or object and a valid path that takes the observer away before returning; 6) \textit{Offscreen Evolution} requires an endogenous process---for example, traffic, smoke, or flowing water---that should continue while temporarily unobserved. The assignment is therefore constrained by scene affordances: exploratory transitions are preferentially assigned to scenes with routes and spatial landmarks, while intentional transitions require scenes with clearly visible targets.

\textbf{Agentic Case Authoring.} With the sampled scene taxonomy and probe family as world metadata, we leverage a series of agents to make case construction efficient at scale while retaining quality control:
\begin{itemize}
\item \textbf{Image Generator.} The sampled metadata is converted into a structured image-generation prompt, which is fed to a generation model to create the initial image. This image serves as the initial observation for the world model.
\item \textbf{Image-grounded Planner.} The planner grounds a concrete action in the visible world. Taking the generated image as input, it specifies the actions for the case, including the text instruction, camera trajectory, control sequence, rollout plan, and physical parameter conditions. It ensures the action is meaningful, diagnosable, and supported by observable evidence via image-grounded reasoning. The planner cannot change the assigned probe family or introduce entities absent from the image.
\item \textbf{Case Validator.} A separate validation agent audits the proposed image--action pair. It verifies that the target is visible and identifiable, the action is feasible in the depicted world, the expected outcome is sufficiently specific, and the rollout will contain evidence adequate for the assigned probe. Taking revisit and offscreen cases as examples, it checks that the return path or hidden process is well defined. Candidates that fail the validity gate are returned to the sampler for resampling or regeneration rather than being silently retained. This validation loop is valuable not only for quality control but also for scalability: it concentrates expensive reasoning on ambiguous candidates while allowing the majority of case authoring steps to be automated.
\end{itemize}

\begin{figure*}[t]
    \centering
    \includegraphics[width=\textwidth]{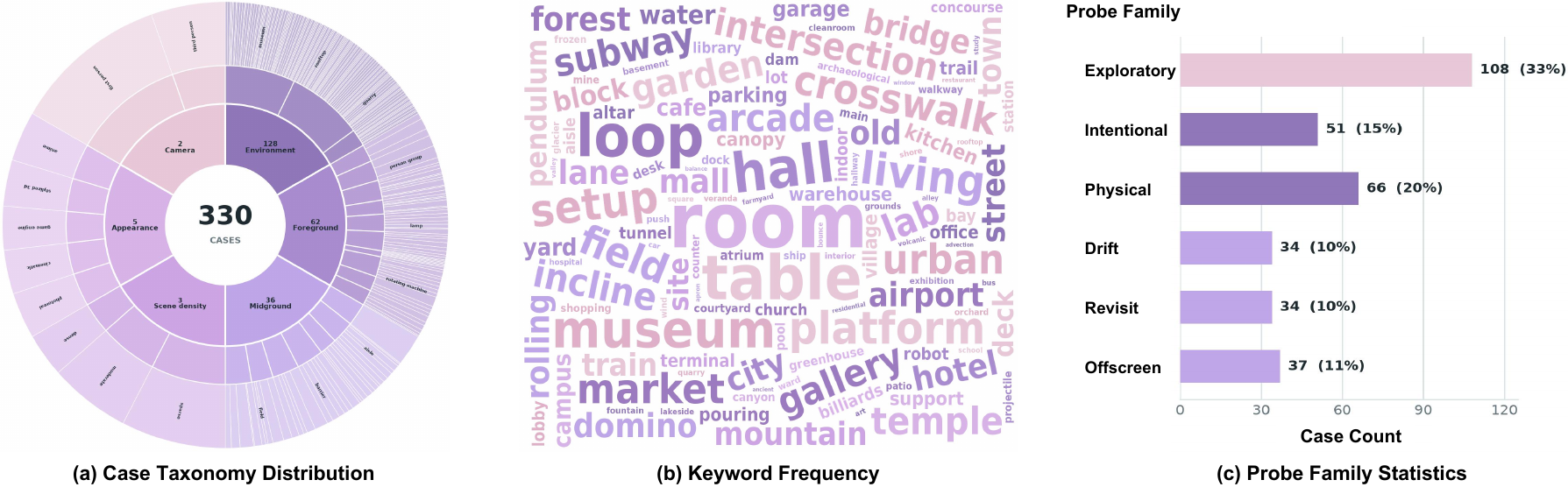}
    \caption{\textbf{Statistics of the HarnessEval-W cases.} (a) Case taxonomy distribution in HarnessEval-W. (b) Keyword frequency over scene descriptions. (c) Probe family distribution. 
}
    \label{fig:case-coverage}
\end{figure*}

\textbf{Benchmark Composition.} Following the construction pipeline, we instantiate a benchmark of 330 cases spanning diverse initial worlds and interaction patterns. As shown in Figure~\ref{fig:case-coverage}, the released cases cover a broad range of environments, foreground entities, spatial layouts, scene densities, visual appearances, and perspectives. The keyword cloud summarizes the dominant scene content, while the horizon panel shows how the six probe families distribute across short intervention cases and longer persistence-style rollouts.

Beyond scene diversity, \ourmodel covers six complementary probe families with different interaction horizons: transition-oriented cases range from direct state or physical interventions to short exploratory rollouts, while persistence-oriented cases require longer trajectories to assess drift resistance, revisit consistency, and offscreen evolution. Together, these cases evaluate both immediate world-state transitions and the consistency of world dynamics over time.

%% file: sections/5_experiments.tex
\section{Experiments and Analysis}
\label{sec:experiments}

In this section, we first introduce the experimental setup (Section \ref{sec:experimental_setup}) and report how current world models rank under \ourmodel, together with the qualitative reasoning traces behind the scores (Section~\ref{sec:main_leaderboard}). We then evaluate the evaluator itself, including the alignment with human preferences, advantage over existing evaluation benchmarks, and evaluation robustness (Section~\ref{sec:evaluator_ranking_robustness}). Finally, we analyze the correlations among evaluation axes and how capabilities shift under finetuning (Section~\ref{sec:analysis}).

\input{tables/tab_leaderboard.tex}

\subsection{Experimental Setup}
\label{sec:experimental_setup}

\textbf{Evaluated World Models.}
We evaluate 18 representative world models, spanning general-purpose video generative models and models designed for interactive world simulation. For each model, we use the released checkpoint or the official API and retain its native conditioning interface. 

\textbf{Evaluation cases.}
All models are evaluated on the same frozen set of 330 \ourmodel cases. Each case specifies an initial observation, an interaction, and the world-state question to be evaluated. Since the models expose different conditioning mechanisms, we translate each interaction into the model's native input form---a text instruction, camera trajectory, or control sequence---preserving the case intent and evaluation target while allowing each model to operate through its intended interface.

\textbf{Evaluation Process.}
We evaluate every generated rollout with the \ourmodel evaluation harness shown in Figure~\ref{fig:pipeline}. For each evaluation case, the harness interprets the assigned world-state question, selects a primary evaluation skill, and composes the supporting skills needed for that specific case. After that, each skill decomposes the question into measurable subproblems, delegates them to specialized sub-agents or tools, and validates the collected evidence before producing a score. All sub-agents use a VLM with the same backend, temperature, and frame-sampling configuration.

\textbf{Evaluation Metrics.}
Following Table~\ref{tab:evaluation_axes}, we design our 8 metrics in correspondence with 8 evaluation settings, \textit{i.e.}, Render Quality (Obs-R), Physical Observation Quality (Obs-P), Exploratory Transition Correctness (Trans-E), Intentional Transition Correctness (Trans-I), Physical Transition Correctness (Trans-P), Drift Resistance (Pers-D), Revisit Consistency (Pers-R) and Offscreen Evolution (Pers-O). Each metric originally ranges from 0 to 1, and we use a normalized final score ranging from 1 to 100 for clear representation. The observation-related metrics (\textit{i.e.}, Obs-R and Obs-P) are evaluated for every case and averaged across all 330 cases, while other metrics are only averaged over the subset of cases with the corresponding probe family. The overall score is computed by averaging the case-level scores.

\subsection{Main Results}
\label{sec:main_leaderboard}
We first demonstrate our \ourmodel leaderboard in Table~\ref{tab:harnessevalw-main}. All 18 models are evaluated on the same 330 cases. We cover three types of world models: general-purpose video generators driven by a text instruction, and autoregressive video models driven by camera pose or by native action controls. As all types of models are evaluated on the same questions with the same cases, we report them in a single table and group them according to the conditioning interface.

Quantitatively, Seedance 2.0~\cite{seedance2026v2} ranks first with an Overall score of $75.5$, followed by Wan 2.7~\cite{wan2025wan} ($75.0$), Kling 3.0~\cite{kuaishou2026kling3} ($74.4$), and MiniMax H3~\cite{minimax2026h3} ($74.3$), where all four models are text-driven general-purpose video generators. We attribute their advantage to large-scale text-conditioned training, which supplies strong grounding and predictive ability to interpret the consequences of an action. Such ability is precisely what Intentional and Physical Transition demand. 

We further examine the scores across 3 evaluation axes and 8 detailed metrics. We find that different models lead in different metrics. For example, Wan 2.7~\cite{wan2025wan} takes the leading performance in Intentional and Physical Transition Correctness, while Seedance 2.0~\cite{seedance2026v2}, HY-WorldPlay 1.5~\cite{sun2025worldplay} and SANA-WM~\cite{zhu2026sanawm} outperforms other models in Drift Resistance, Revisit Consistency and Offscreen Evolution, respectively. This strength diversity mainly comes from differences in the training data distributions of these models. 

\begin{figure*}[t]
    \centering
    \includegraphics[width=\linewidth]{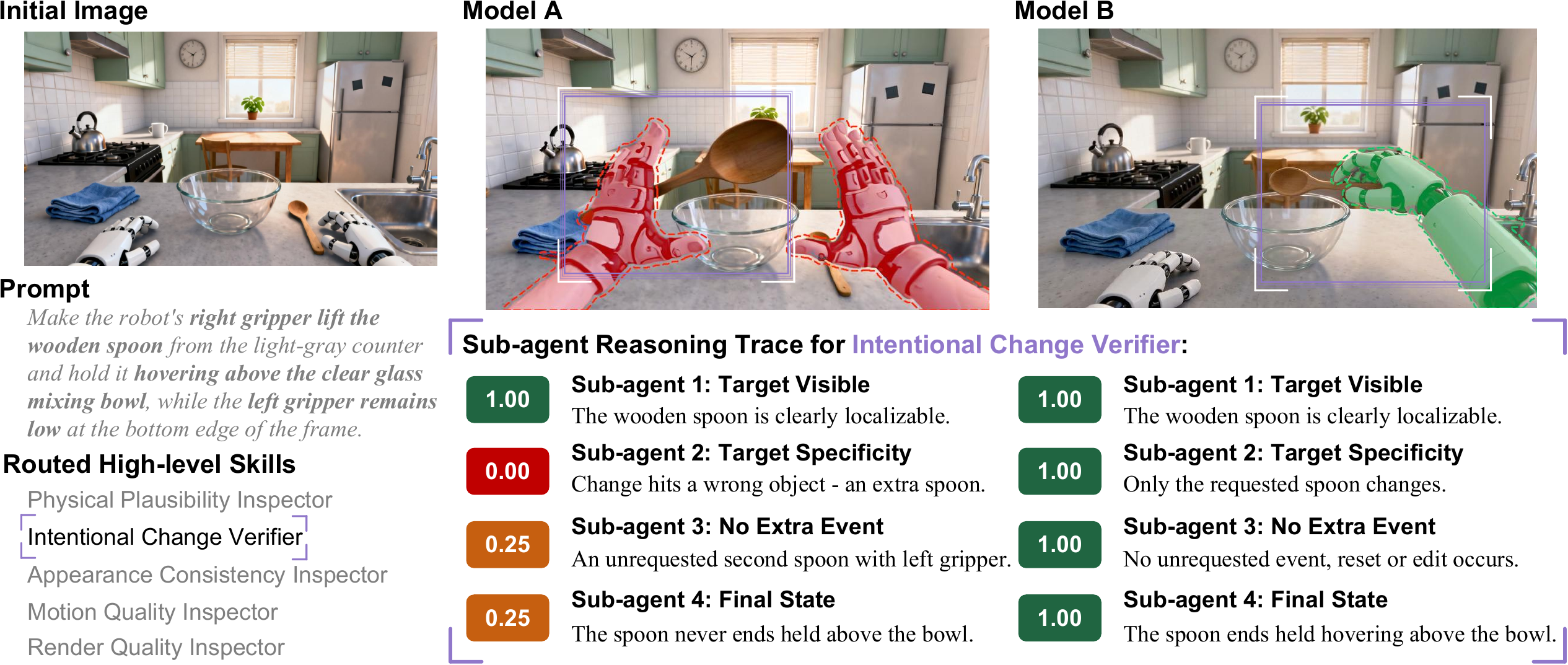}
    \caption{
    \textbf{Complete reasoning traces produced by \ourmodel.} Each card records the full trajectory of one evaluation: the case specification, the generated rollout, the selected evaluation route with evidence-grounded reasons for each skill. For the long-horizon navigation case (left), the planner selects drift analysis as the core skill; for the intentional state-change case (right), it selects the intentional-change verifier while skipping persistence- and physics-specific skills.}
    \label{fig:qualitative_case_cards}
\end{figure*}

We then qualitatively inspect the complete reasoning traces produced for each evaluation case and model. Each trace records the case specification, the generated rollout, the LLM planner's selected evaluation route, the skill-level measurements, the aggregation result, and the validator's decisions, making every evaluation auditable end to end. Figure~\ref{fig:qualitative_case_cards} illustrates two distinct planning routes. For the long-horizon navigation case, the planner selects drift analysis as the core skill, with render-quality and motion evidence providing complementary signals; for the intentional state-change case, it instead selects the intentional-change verifier while skipping persistence- and physics-specific skills. In both cases, the validator audits the proposed route and adds the required evidence gates before execution. 

\begin{figure}[t]
    \centering
    \includegraphics[width=\textwidth]{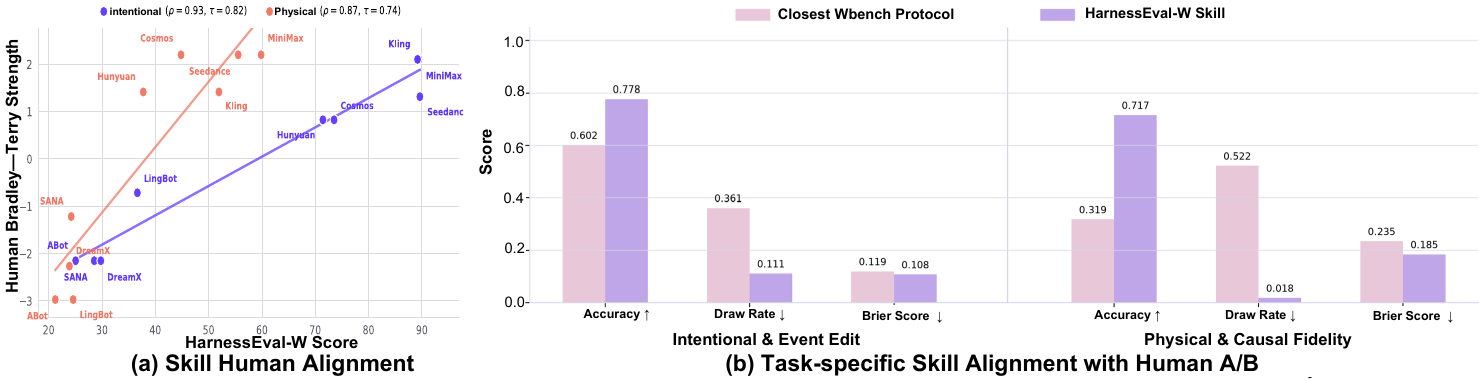}
    \caption{\textbf{Human alignment of \ourmodel and its comparison with WBench.}
(a) Model-level human alignment: each point is one of the evaluated models; we fit a linear curve for both  Intentional and Physical Transition. (b) Controlled comparison with the closest WBench protocols on the same data: we report pairwise accuracy (higher is better), draw rate, and Brier score (both lower is better).}
    \label{fig:human_and_task_specific_alignment}
\end{figure}

\subsection{Evaluating the Evaluator: Human Alignment, Robustness and Comparison}
\label{sec:evaluator_ranking_robustness}

We evaluate the performance of \ourmodel itself in this section. Specifically, we assess \ourmodel from three angles: whether its scores align with human preference, how it compares with other benchmarks, and its robustness under different VLMs. We use two aspects for evaluation: Intentional and Physical Transition, the two settings that most directly require semantic and causal interpretation of an intervention and are therefore the hardest to evaluate automatically.

\textbf{Human Alignment.} 
We first conduct a human study to establish a reference ordering. Annotators are shown pairs of rollouts generated by two different models for the same case and asked to choose the better one; we collect 5000 such A/B judgments across nine representative models. The pairwise choices are then aggregated into a per-model strength score with a Bradley--Terry model.

As reported in Figure~\ref{fig:human_and_task_specific_alignment}(a), \ourmodel closely aligns with the resulting human ordering on both Intentional and Physical Transition, achieving a Spearman rank correlation of $\rho = \text{0.93}$ (Kendall $\tau = \text{0.82}$) on Intentional Transition and $\rho = \text{0.87}$ ($\tau = \text{0.74}$) on Physical Transition.

\textbf{Comparison with WBench~\cite{ying2026wbench}.}
We next compare \ourmodel against the closest existing evaluation protocols, drawn from WBench~\cite{ying2026wbench}: Event Edit for Intentional and Causal Fidelity for Physical. Both protocols are re-run on the same videos with the same GPT-5.5 backend, temperature, and frame sampling as our benchmark, so that only the evaluation protocol differs. Event Edit asks five binary questions about the video---whether the scene changes, whether the event is recognizable, whether it concludes, whether key details are correct, and whether unrelated anomalies appear---and sums the answers; Causal Fidelity compresses the physics and causal consistency of an entire rollout video into a single $0$--$3$ score. We compare the evaluation results from our benchmark and WBench, and report three metrics in Figure~\ref{fig:human_and_task_specific_alignment}(b): \emph{pairwise accuracy}, whether the evaluator prefers the same model as the human raters in each A/B pair; \emph{draw rate}, the percentage of pairs to which the evaluator assigns identical scores and therefore fails to separate; and \emph{Brier score}, the mean squared error between the win probability implied by the evaluator's scores and the human choice. Across all three metrics, \ourmodel aligns substantially better with human judgment: on Physical, it raises pairwise accuracy from $31.9\%$ to $71.7\%$ while cutting the draw rate from $52.2\%$ to $1.8\%$; on Intentional, it raises accuracy from $60.2\%$ to $77.8\%$ with draws falling from $36.1\%$ to $11.1\%$; and it yields the lower Brier score in both settings. These results support the central design choice of \ourmodel: decomposing one global judgment into many separately grounded sub-questions produces evaluation that is both more discriminative and more human-aligned.

\textbf{Robustness of \ourmodel.} Finally, we evaluate the robustness of the evaluator itself by running GPT multiple times with the same GPT-5.5 backend at temperature $0$. We then fit a linear score-to-human curve for each round, and apply the same procedure to the WBench protocol (Figure~\ref{fig:evaluator_robustness}). 
\ourmodel produces consistent evaluation across different rounds: the slopes of fitted curves stay within $9.6$--$10.8$, the correlations with human strength within $0.928$--$0.964$, and the envelope of the three fits spans only $0.33$ Bradley--Terry units. WBench, by contrast, is far less stable: its slope nearly doubles across rounds ($11.2$ to $21.0$), its correlation ranges between $0.646$ and $0.780$, and its envelope spans $1.61$ units---$4.9\times$ wider than ours. \ourmodel therefore not only aligns with human judgment but also returns the evaluation consistently under repetitions, which is closer to an ideal benchmark for world models.

\begin{table}[t]
\centering

\begin{minipage}[t]{0.545\textwidth}
    \vspace{0pt}
    \centering
    \scriptsize

    \includegraphics[width=\linewidth]
        {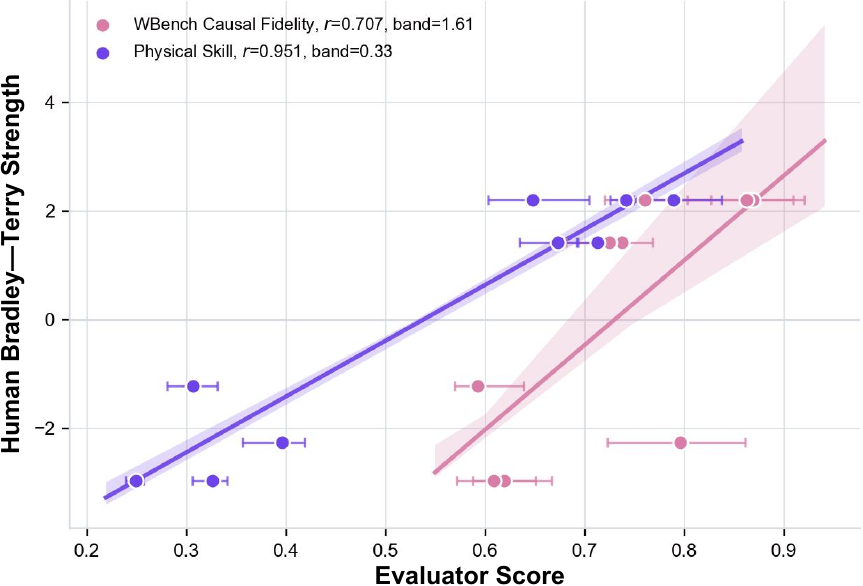}
    \vspace{0.4em}

    \captionof{figure}{
        \textbf{Robustness of evaluation.}
        We run the benchmark three times and fit a linear curve for each run.
        The shaded band is the envelope of the three fits.
        The envelope of \ourmodel is $4.9\times$ narrower than that of WBench.
    }
    \label{fig:evaluator_robustness}
\end{minipage}\hfill%
\begin{minipage}[t]{0.435\textwidth}
    \vspace{0pt}
    \centering
    \scriptsize

    \includegraphics[width=\linewidth]
        {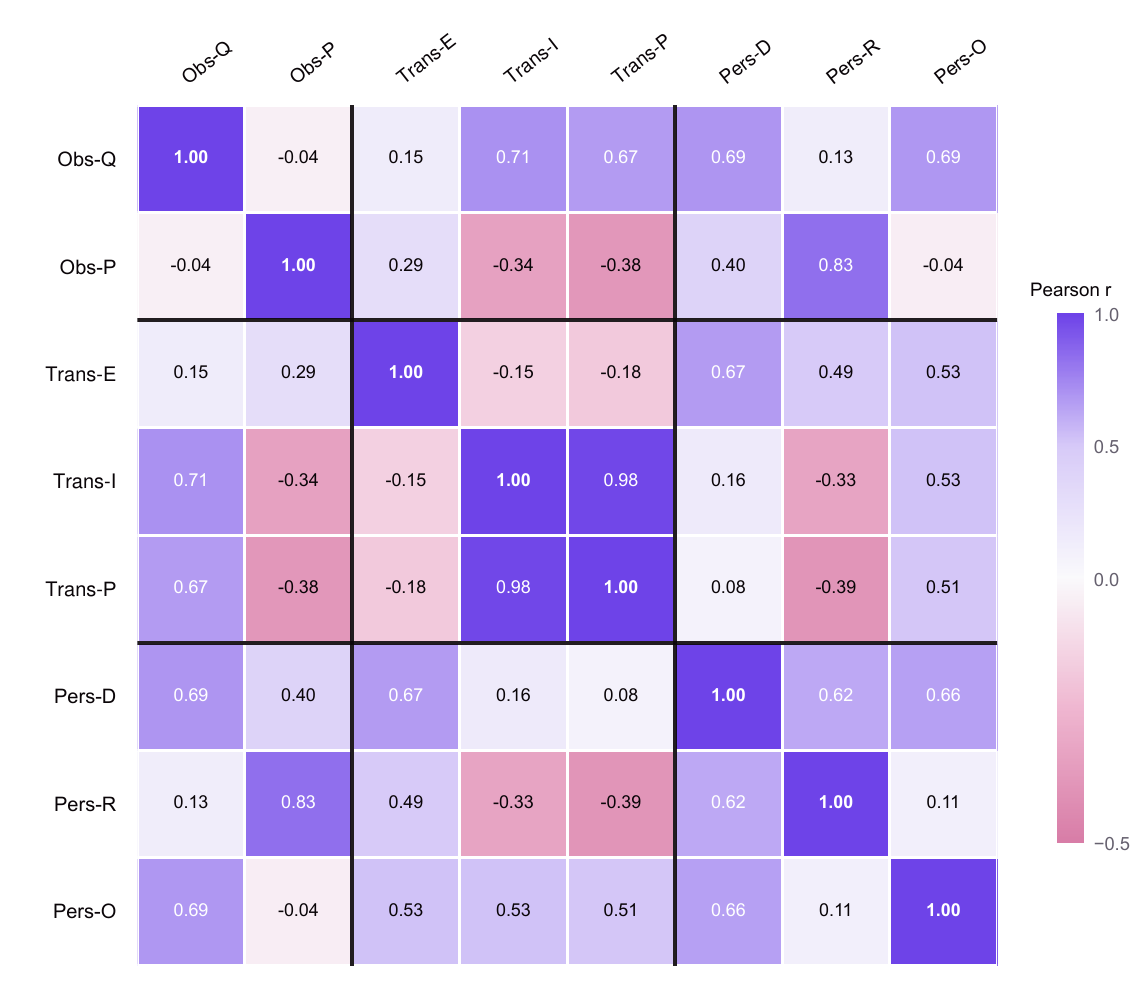}
    \vspace{0.4em}

    \captionof{figure}{
        \textbf{Correlations among evaluation axes.} Each cell reports the Pearson correlation between two axes. The results are reported across all 18 models.
    }
    \label{fig:capability_relationships}
\end{minipage}

\end{table}

\subsection{Further Analysis}
\label{sec:analysis}
We further analyze the benchmark results from two perspectives: the correlations among evaluation settings, and how capabilities shift when a text-to-video generator is fine-tuned into an action-conditioned video model.

\subsubsection{Correlations among Different Evaluation Axes}
\label{sec:capability_relationships}

To understand how the evaluated capabilities relate to one another, we compute pairwise Pearson correlations across the eight evaluated axes over all 18 models and report them in Figure~\ref{fig:capability_relationships}. Render Quality has little correlation with Physical Observation ($r=-0.04$). Intentional and Physical Transition are strongly coupled ($r=0.98$): both require the model to understand the semantics of a specified intervention and generate its corresponding consequences. By contrast, Exploratory Transition is nearly unrelated to either ($r=-0.15$ and $r=-0.18$): producing a coherent exploratory continuation does not imply the semantic and physical grounding required to execute a commanded interaction.

\subsubsection{Capability Shifts under Finetuning}
\label{sec:post_training_retention}
\begin{figure}[t]
    \centering
    \includegraphics[width=0.96\linewidth]{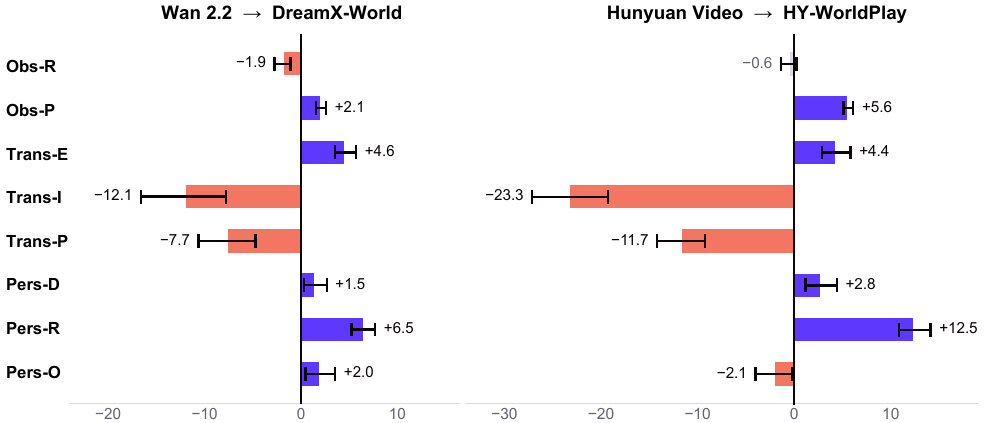}
    \caption{
\textbf{Capability shifts under fine-tuning.} For each pair (Wan 2.2 $\rightarrow$ DreamX-World and HunyuanVideo 1.5 $\rightarrow$ HY-WorldPlay 1.5), we report the per-axis difference $\Delta = S_{\text{fine-tuned}} - S_{\text{original}}$.}
    \label{fig:lineage_paired_deltas}
\end{figure}

We examine how capabilities shift when a text-to-video model is fine-tuned into an action-conditioned video model, and report the results in Figure~\ref{fig:lineage_paired_deltas}. We compare two pairs of models: Wan 2.2 fine-tuned into DreamX-World, and HunyuanVideo 1.5 fine-tuned into HY-WorldPlay 1.5. For each pair, we compute the difference in all the axes $\Delta = S_{\text{fine-tuned}} - S_{\text{original}}$. 
DreamX-World gains $4.8$ points in Exploratory, $7.8$ in Revisit, and $3.5$ in Offscreen, while losing $11.9$ in Intentional and $7.2$ in Physical; HY-WorldPlay gains $8.4$ points in Revisit while losing $24.2$ in Intentional and $11.2$ in Physical. Both pairs exhibit the same trend: the fine-tuned model becomes better at revisiting places it has already seen, and worse at handling physical interactions. We hypothesize that this shift might stem from the fine-tuning data, which emphasizes exploration-style trajectories over commanded interventions.

%% file: tables/tab_leaderboard.tex
\begin{table*}[t]
\centering
\footnotesize
\setlength{\tabcolsep}{1.2pt}
\renewcommand{\arraystretch}{1.18}

\definecolor{LeaderboardFirst}{HTML}{CEC2E2}
\definecolor{LeaderboardSecond}{HTML}{E7D7F1}
\definecolor{LeaderboardThird}{HTML}{EFE2EA}
\newcommand{\leaderboardscore}[3][white]{%
  \begingroup
  \setlength{\fboxsep}{0pt}%
  \colorbox{#1}{%
    \parbox[c][2.25em][c]{\linewidth}{\centering
      \shortstack[c]{\strut #2\\[-2.2pt]
        {\fontsize{5.2}{5.8}\selectfont
         \textcolor{MirrosMuted}{\##3}}}}}%
  \endgroup
}
\newcommand{\leaderboardfirst}[2]{%
  \leaderboardscore[LeaderboardFirst]{\bfseries #1}{#2}}
\newcommand{\leaderboardsecond}[2]{%
  \leaderboardscore[LeaderboardSecond]{#1}{#2}}
\newcommand{\leaderboardthird}[2]{%
  \leaderboardscore[LeaderboardThird]{#1}{#2}}
\newcommand{\leaderboardhead}[1]{%
  \scriptsize\bfseries #1}
\newcommand{\leaderboardinterface}[1]{%
  \scriptsize\textcolor{MirrosMuted}{#1}}
\newcommand{\leaderboarddivider}{%
  \noalign{\vskip 2.2pt\color{MirrosLine}\hrule height 0.35pt\vskip 2.2pt}}

\caption{\textbf{HarnessEval-W main leaderboard.} All models are evaluated
on the same case set and grouped by conditioning interface. Overall is the
arithmetic average of the 330 cases. The gray label below each score gives
its column-wise rank across all 18 models. Purple backgrounds mark first, second, and third place from
darkest to lightest. Close-source methods are annotated with $^*$.}
\label{tab:harnessevalw-main}
\begin{tabularx}{\textwidth}{@{}ll@{\hspace{2.4pt}}*{9}{>{\centering\arraybackslash}X@{}}}
\toprule
\multirow{2}{*}{\textbf{Model}} & \multirow{2}{*}{\textbf{Interface}} & \multicolumn{2}{c}{\textbf{Observation }}
& \multicolumn{3}{c}{\textbf{Transition }}
& \multicolumn{3}{c}{\textbf{ Persistence}}
& \\
\cmidrule(lr){3-4}\cmidrule(lr){5-7}\cmidrule(lr){8-10}
&
& \leaderboardhead{Obs-Q} & \leaderboardhead{Obs-P} & \leaderboardhead{Trans-E}
& \leaderboardhead{Trans-I} & \leaderboardhead{Trans-P} & \leaderboardhead{Pers-D}
& \leaderboardhead{Pers-R} & \leaderboardhead{Pers-O}
& \leaderboardhead{Overall} \\
\midrule

Seedance 2.0$^*$~\cite{seedance2026v2}
& \leaderboardinterface{Prompt I2V}
& \leaderboardsecond{83.6}{2}
& \leaderboardscore{61.8}{7}
& \leaderboardscore{80.2}{7}
& \leaderboardthird{81.8}{3}
& \leaderboardscore{63.5}{4}
& \leaderboardfirst{79.8}{1}
& \leaderboardscore{76.8}{5}
& \leaderboardsecond{68.9}{2}
& \leaderboardfirst{75.5}{1} \\
Wan 2.7$^*$~\cite{wan2025wan}
& \leaderboardinterface{Prompt I2V}
& \leaderboardscore{80.9}{8}
& \leaderboardscore{58.8}{16}
& \leaderboardscore{78.7}{9}
& \leaderboardfirst{83.6}{1}
& \leaderboardfirst{71.1}{1}
& \leaderboardscore{74.3}{16}
& \leaderboardscore{68.5}{15}
& \leaderboardscore{65.9}{8}
& \leaderboardsecond{75.0}{2} \\
Kling 3.0$^*$~\cite{kuaishou2026kling3}
& \leaderboardinterface{Prompt I2V}
& \leaderboardscore{82.1}{4}
& \leaderboardscore{60.6}{12}
& \leaderboardscore{79.1}{8}
& \leaderboardsecond{82.6}{2}
& \leaderboardscore{63.2}{5}
& \leaderboardscore{77.2}{9}
& \leaderboardscore{75.4}{7}
& \leaderboardscore{66.2}{6}
& \leaderboardthird{74.4}{3} \\
MiniMax H3~\cite{minimax2026h3}
& \leaderboardinterface{Prompt I2V}
& \leaderboardscore{81.5}{6}
& \leaderboardscore{61.3}{9}
& \leaderboardscore{77.7}{11}
& \leaderboardscore{81.7}{4}
& \leaderboardsecond{66.9}{2}
& \leaderboardscore{77.0}{11}
& \leaderboardscore{72.3}{9}
& \leaderboardscore{67.2}{4}
& \leaderboardscore{74.3}{4} \\
Grok Imagine 1.5$^*$~\cite{xai2026grokimaginevideo15}
& \leaderboardinterface{Prompt I2V}
& \leaderboardfirst{85.1}{1}
& \leaderboardscore{60.6}{12}
& \leaderboardscore{76.4}{14}
& \leaderboardscore{80.2}{5}
& \leaderboardthird{66.7}{3}
& \leaderboardscore{79.1}{4}
& \leaderboardscore{70.9}{11}
& \leaderboardscore{64.5}{10}
& \leaderboardscore{73.4}{5} \\
FLUX 3$^*$~\cite{blackforestlabs2026flux3video}
& \leaderboardinterface{Prompt I2V}
& \leaderboardscore{81.7}{5}
& \leaderboardscore{61.6}{8}
& \leaderboardscore{76.4}{14}
& \leaderboardscore{79.0}{6}
& \leaderboardscore{63.2}{5}
& \leaderboardscore{77.1}{10}
& \leaderboardscore{70.2}{13}
& \leaderboardthird{67.4}{3}
& \leaderboardscore{72.2}{6} \\
Cosmos3-Super~\cite{nvidia2025cosmosplatform}
& \leaderboardinterface{Prompt I2V}
& \leaderboardthird{83.5}{3}
& \leaderboardscore{61.0}{10}
& \leaderboardscore{77.9}{10}
& \leaderboardscore{75.1}{7}
& \leaderboardscore{60.2}{7}
& \leaderboardscore{77.6}{6}
& \leaderboardscore{70.6}{12}
& \leaderboardscore{66.8}{5}
& \leaderboardscore{71.9}{7} \\
HunyuanVideo 1.5~\cite{wu2025hunyuanvideo15}
& \leaderboardinterface{Prompt I2V}
& \leaderboardscore{80.0}{11}
& \leaderboardscore{59.0}{15}
& \leaderboardscore{77.6}{12}
& \leaderboardscore{73.2}{8}
& \leaderboardscore{57.2}{8}
& \leaderboardscore{76.5}{12}
& \leaderboardscore{69.5}{14}
& \leaderboardscore{63.2}{12}
& \leaderboardscore{70.3}{8} \\
Wan 2.2~\cite{wan2025wan}
& \leaderboardinterface{Prompt I2V}
& \leaderboardscore{80.4}{10}
& \leaderboardscore{58.6}{17}
& \leaderboardscore{77.2}{13}
& \leaderboardscore{62.0}{9}
& \leaderboardscore{55.1}{9}
& \leaderboardscore{76.0}{14}
& \leaderboardscore{66.5}{17}
& \leaderboardscore{63.4}{11}
& \leaderboardscore{67.7}{11} \\
LTX-2.3~\cite{ha2026ltx2}
& \leaderboardinterface{Prompt I2V}
& \leaderboardscore{78.4}{16}
& \leaderboardscore{54.5}{18}
& \leaderboardscore{73.8}{16}
& \leaderboardscore{60.4}{10}
& \leaderboardscore{53.3}{10}
& \leaderboardscore{72.4}{17}
& \leaderboardscore{63.3}{18}
& \leaderboardscore{57.7}{15}
& \leaderboardscore{64.6}{16} \\
\leaderboarddivider

SANA-WM~\cite{zhu2026sanawm}
& \leaderboardinterface{Native action}
& \leaderboardscore{81.0}{7}
& \leaderboardscore{62.3}{6}
& \leaderboardsecond{82.5}{2}
& \leaderboardscore{50.6}{13}
& \leaderboardscore{47.6}{13}
& \leaderboardscore{78.9}{5}
& \leaderboardthird{78.8}{3}
& \leaderboardfirst{72.3}{1}
& \leaderboardscore{68.7}{10} \\
ABot-World~\cite{jiang2026abotworld}
& \leaderboardinterface{Native action}
& \leaderboardscore{77.7}{17}
& \leaderboardscore{60.8}{11}
& \leaderboardfirst{83.5}{1}
& \leaderboardscore{49.0}{16}
& \leaderboardscore{45.7}{15}
& \leaderboardscore{76.3}{13}
& \leaderboardscore{72.0}{10}
& \leaderboardscore{60.8}{14}
& \leaderboardscore{66.1}{14} \\
DreamX-World~\cite{dreamx2026world}
& \leaderboardinterface{Native action}
& \leaderboardscore{78.5}{15}
& \leaderboardscore{60.6}{12}
& \leaderboardscore{81.8}{4}
& \leaderboardscore{50.0}{14}
& \leaderboardscore{47.4}{14}
& \leaderboardscore{77.5}{8}
& \leaderboardscore{73.1}{8}
& \leaderboardscore{65.4}{9}
& \leaderboardscore{66.8}{13} \\
\leaderboarddivider
LingBot World v2~\cite{gao2026lingbotworld}
& \leaderboardinterface{Camera pose}
& \leaderboardscore{80.5}{9}
& \leaderboardthird{63.3}{3}
& \leaderboardscore{81.5}{5}
& \leaderboardscore{56.8}{11}
& \leaderboardscore{49.8}{11}
& \leaderboardsecond{79.5}{2}
& \leaderboardscore{75.8}{6}
& \leaderboardscore{66.1}{7}
& \leaderboardscore{68.8}{9} \\
Lyra 2~\cite{shen2026lyra2}
& \leaderboardinterface{Camera pose}
& \leaderboardscore{79.5}{12}
& \leaderboardsecond{64.0}{2}
& \leaderboardscore{80.7}{6}
& \leaderboardscore{48.9}{17}
& \leaderboardscore{45.1}{18}
& \leaderboardscore{77.6}{6}
& \leaderboardsecond{79.7}{2}
& \leaderboardscore{56.4}{16}
& \leaderboardscore{65.5}{15} \\
Fantasy-World~\cite{dai2025fantasyworld}
& \leaderboardinterface{Camera pose}
& \leaderboardscore{74.1}{18}
& \leaderboardscore{62.7}{5}
& \leaderboardscore{73.2}{17}
& \leaderboardscore{52.4}{12}
& \leaderboardscore{48.9}{12}
& \leaderboardscore{69.9}{18}
& \leaderboardscore{68.4}{16}
& \leaderboardscore{53.8}{18}
& \leaderboardscore{62.1}{17} \\
HY-WorldPlay 1.5~\cite{sun2025worldplay}
& \leaderboardinterface{Camera pose}
& \leaderboardscore{79.4}{13}
& \leaderboardfirst{64.6}{1}
& \leaderboardthird{82.0}{3}
& \leaderboardscore{49.9}{15}
& \leaderboardscore{45.5}{17}
& \leaderboardthird{79.3}{3}
& \leaderboardfirst{81.9}{1}
& \leaderboardscore{61.1}{13}
& \leaderboardscore{67.1}{12} \\
InSpatio-World~\cite{shen2026inspatioworld}
& \leaderboardinterface{Camera pose}
& \leaderboardscore{79.0}{14}
& \leaderboardscore{63.0}{4}
& \leaderboardscore{70.7}{18}
& \leaderboardscore{48.6}{18}
& \leaderboardscore{45.6}{16}
& \leaderboardscore{74.5}{15}
& \leaderboardscore{77.0}{4}
& \leaderboardscore{54.0}{17}
& \leaderboardscore{61.4}{18} \\

\bottomrule
\end{tabularx}
\end{table*}

%% file: sections/6_future_work.tex
\section{Future Work: Toward Self-Evolving Harness for Evaluation}
Evaluation is never a static rubric. It evolves in our evolving world, mirroring a broader shift toward harness engineering and recursive self-improvement~\citep{weng2026harness,chen2026rsi,anthropic2026rsi}.
\ourmodel is only the start of this agentic benchmark paradigm; we identify three directions in which the evaluation harness should evolve alongside the worlds it measures.

\textbf{Test-Time Scaling for Agentic Benchmarks.}
As LLMs generate increasingly complex reasoning traces and leverage external tools to solve long-horizon tasks, the evaluation process must scale accordingly. The same test-time-compute paradigm that makes models more capable~\citep{snell2024scaling,muennighoff2025s1} applies equally to the evaluator: within our harness, additional compute buys finer skill decomposition, deeper sub-agent search over the rollout, multi-step verification, and repeated re-verification of the collected evidence, each of which makes a verdict strictly more complete. 
This paradigm shift ensures the evaluator is always as capable as the model being tested.

\textbf{Scaling Skill Libraries.}
Future world models will scale up and serve a vastly broader range of scenarios, generating highly complex environments with fine-grained physical fidelity, and the evaluator must scale its capabilities accordingly. We therefore need to continuously grow a comprehensive skill library. Each skill encodes human knowledge about how to evaluate a specific a scenario---what evidence settles a question, and which tool grounds it---so that scaling the library accumulates evaluation knowledge rather than re-deriving a bespoke rubric for every new benchmark, in the same spirit as agents that grow reusable libraries of executable skills~\citep{wang2023voyager,zhang2026rewardharness}. As models generate more intricate physics and more diverse scenes, the evaluator dynamically retrieves and composes specialized skills from this ever-growing library to accurately assess physical understanding across generated worlds.

\textbf{Recursively Self-Improving Agentic Benchmarks.}
When the system encounters out-of-distribution scenarios while evaluating frontier models---when routing finds no skill that can legitimately answer a case---this failure is a measurement of the evaluator's own limitation. We intend to make such skill gaps explicit and actionable: through external skill expansion,  or self-driven exploration~\citep{hu2024adas,zhang2025darwingodel,wang2026agenticeval}, the evaluation agent acquires the missing capabilities and writes them back into its skill library, creating a recursively self-improving evaluation loop that never becomes obsolete. An evaluator that can extend itself into worlds it has never seen is what makes self-improvement possible beyond the known world~\citep{mirros2026physicalrsi}.

%% file: sections/7_conclusion.tex
\section{Conclusion}
In this paper, we introduced \ourmodel, an agentified evaluation pipeline that brings the harness paradigm from the LLM ecosystem to world model benchmarking. Rather than applying a fixed rubric, \ourmodel interprets the context of each evaluation case, decomposes the evaluation into measurable sub-questions answered by specialized sub-agents, and aggregates the validated evidence into a transparent reasoning trace, so that every score can be examined and verified. Across 330 evaluation cases and 18 representative world models, its judgments closely align with human preferences, faithfully reflecting model performance. We open-source \ourmodel as a living benchmark: an executable agentic system that grows new skills and evaluation cases as world models evolve. We invite the broader community to contribute to this agentic benchmark workflow together.

%% file: sections/X_appendix.tex
\section{Evaluated Models}
\label{app:evaluated_models}

\subsection{Model Adapter}
\label{app:model_adapter}

Following WBench~\cite{ying2026wbench}, the model adapter translates each case's
fixed initial observation and action semantics into the model's native
conditioning interface. Prompt-I2V models receive the initial image and a
natural-language instruction, camera-pose models receive a 6-DoF camera
trajectory, and native-action models receive the corresponding keyboard or
control sequence. The generated rollout is converted to a common video format
and passed to the same evaluator. Thus, the adapter changes only the
representation of an action, not the action being evaluated.

\begin{table}[!ht]
\centering
\scriptsize
\setlength{\tabcolsep}{5pt}
\renewcommand{\arraystretch}{1.28}
\begin{tabularx}{\textwidth}{m{0.20\linewidth}m{0.27\linewidth}m{\dimexpr\linewidth-0.47\linewidth-6\tabcolsep\relax}}
\toprule
\textbf{Model family} & \textbf{Fixed case input} & \textbf{Model-facing representation} \\
\midrule
Prompt-I2V & Initial observation and action specification & Initial image plus a natural-language prompt describing the same action. \\
Camera-pose & Initial observation and navigation specification & Initial image plus a 6-DoF camera or subject-motion trajectory. \\
Native-action & Initial observation and control specification & Initial image plus the model's native keyboard or action sequence. \\
\bottomrule
\end{tabularx}
\caption{\textbf{Model adapter contracts.} All families use the same case semantics; the adapter changes only the model-facing representation.}
\label{tab:app_model_adapter}
\end{table}
\FloatBarrier

\subsection{Model Configuration}
\label{app:evaluated_model_configuration}

\begin{table}[!ht]
\centering
\scriptsize
\setlength{\tabcolsep}{4pt}
\renewcommand{\arraystretch}{1.28}
\begin{tabularx}{\textwidth}{m{0.22\linewidth}m{0.08\linewidth}m{0.15\linewidth}m{0.07\linewidth}m{0.13\linewidth}m{0.11\linewidth}m{\dimexpr\linewidth-0.76\linewidth-14\tabcolsep\relax}}
\toprule
\textbf{Model} & \textbf{Access} & \textbf{Architecture} & \textbf{Params} & \textbf{Resolution} & \textbf{Frames / turn} & \textbf{FPS} \\
\midrule
\multicolumn{7}{l}{\textit{Prompt-I2V models}} \\
Seedance 2.0~\cite{seedance2026v2} & API & -- & -- & $864{\times}496$ & 97 & 24 \\
Wan 2.7~\cite{wan2025wan} & API & -- & -- & $1280{\times}720$ & 60 & 30 \\
Kling 3.0~\cite{kuaishou2026kling3} & API & -- & -- & $1280{\times}720$ & 121 & 24 \\
MiniMax H3~\cite{minimax2026h3} & Open & DiT & 33B & $896{\times}512$ & 39 & 24 \\
Grok Imagine Video 1.5~\cite{xai2026grokimaginevideo15} & API & -- & -- & $848{\times}480$ & 49 & 24 \\
FLUX 3~\cite{blackforestlabs2026flux3video} & API & Multimodal flow & -- & $1280{\times}704$ & 121 & 24 \\
Cosmos3-Super~\cite{nvidia2025cosmosplatform} & Open & DiT & 64B & $832{\times}480$ & 48 & 24 \\
HunyuanVideo 1.5~\cite{wu2025hunyuanvideo15} & Open & DiT & 8B & $848{\times}480$ & 48 & 24 \\
Wan 2.2~\cite{wan2025wan} & Open & DiT & 5B & $1248{\times}704$ & 48 & 24 \\
LTX-2.3~\cite{ha2026ltx2} & Open & DiT & 22B & $768{\times}512$ & 48 & 24 \\
\midrule
\multicolumn{7}{l}{\textit{Camera-pose models}} \\
LingBot World v2~\cite{gao2026lingbotworldv2} & Open & Multi-stage DiT & 14B & $832{\times}464$ & 32 & 16 \\
HY-WorldPlay 1.5~\cite{sun2025worldplay} & Open & Streaming DiT & 8B & $832{\times}480$ & 48 & 24 \\
Fantasy-World~\cite{dai2025fantasyworld} & Open & DiT + 3D head & 14B & $592{\times}336$ & 32 & 16 \\
InSpatio-World~\cite{shen2026inspatioworld} & Open & V2V AR DiT & 1.3B & $832{\times}480$ & 48 & 24 \\
Lyra 2~\cite{shen2026lyra2} & Open & DiT + 3D head & 14B & $832{\times}480$ & 32 & 16 \\
\midrule
\multicolumn{7}{l}{\textit{Native-action models}} \\
SANA-WM~\cite{zhu2026sanawm} & Open & Streaming DiT & 2B & $1280{\times}704$ & 32 & 16 \\
DreamX-World~\cite{dreamx2026world} & Open & AR DiT & 5B & $1280{\times}704$ & 32 & 16 \\
ABot-World~\cite{jiang2026abotworld} & Open & AR DiT & 5B & $832{\times}480$ & 24 & 12 \\
\bottomrule
\end{tabularx}
\caption{\textbf{Overview of the 18 evaluated models.} ``Frames / turn'' is the number of output frames used for one interaction turn; FPS is the output frame rate.}
\label{tab:app_evaluated_models}
\end{table}
\FloatBarrier

The descriptions below summarize the models themselves; benchmark-specific
output and sampling settings are reported only in Table~\ref{tab:app_evaluated_models}.

\paragraph{Prompt-I2V models.}

\paragraph{Seedance 2.0.}
Developed by ByteDance's Seedance team, Seedance 2.0 is a native multimodal
audio-video generator with a unified large-scale architecture. It jointly
accepts text, image, audio, and video references and supports generation and
editing within the same model.~\cite{seedance2026v2}

\paragraph{Wan 2.7.}
Alibaba's Wan team develops Wan as a family of large-scale video foundation
models built on the diffusion-transformer paradigm. Wan 2.7 is the hosted
image-to-video member evaluated here; its serving architecture is not publicly
disclosed.~\cite{wan2025wan}

\paragraph{Kling 3.0.}
Kling 3.0 is Kuaishou Technology's proprietary multimodal video-generation
system. Kling AI presents it as a high-fidelity model centered on physics-aware
motion, subject consistency, and native synchronized audio; its internal network
architecture is not disclosed.~\cite{kuaishou2026kling3}

\paragraph{MiniMax H3.}
MiniMax H3 is MiniMax's open general-purpose multimodal generation model. Its
H3-Omni Transformer, H3-VAE, Contextual Omni Representation, and in-context
regeneration unify understanding and generation across text, image, video, and
audio.~\cite{minimax2026h3}

\paragraph{Grok Imagine Video 1.5.}
Grok Imagine Video 1.5 is xAI's proprietary multimodal video-generation model,
offered through hosted interfaces for text- and image-conditioned creation. xAI
does not publicly disclose its network architecture or parameter count.~\cite{xai2026grokimaginevideo15}

\paragraph{FLUX 3.}
Developed by Black Forest Labs, FLUX 3 is a multimodal foundation model that
jointly learns from images, video, and audio in a unified flow-matching
architecture. Its Self-Flow approach aligns multimodal generation and
understanding within the same backbone, while FLUX 3 Video provides text- and
image-conditioned video generation.~\cite{blackforestlabs2026flux3,blackforestlabs2026flux3video}

\paragraph{Cosmos3-Super.}
Cosmos3-Super belongs to NVIDIA's open Cosmos world foundation model platform
for Physical AI. The platform couples video tokenization with diffusion-based
world foundation models that can be post-trained into application-specific
simulators; Cosmos3-Super is its image-to-video generation variant.~\cite{nvidia2025cosmosplatform}

\paragraph{HunyuanVideo 1.5.}
Developed by Tencent Hunyuan, HunyuanVideo 1.5 is an open video DiT for
text-to-video and image-to-video generation. Its selective and sliding tile
attention, glyph-aware bilingual text encoding, and progressive training target
motion coherence and efficient deployment.~\cite{wu2025hunyuanvideo15}

\paragraph{Wan 2.2.}
Alibaba's Wan 2.2 is an open diffusion-transformer video-generation family. Its
large variants introduce timestep-specialized mixture-of-experts denoisers,
while the compact TI2V branch jointly supports text- and image-conditioned
generation through a high-compression VAE.~\cite{wan2025wan}

\paragraph{LTX-2.3.}
LTX-2.3 is Lightricks' open audiovisual foundation model. Its asymmetric
dual-stream transformer uses separate video and audio streams connected by
bidirectional cross-attention, enabling temporally synchronized video, speech,
ambience, and sound effects from shared conditioning.~\cite{ha2026ltx2}

\paragraph{Camera-pose models.}

\paragraph{LingBot World v2.}
Robbyant's LingBot World v2 is an open interactive video world model built on
Wan2.2. It combines causal pretraining with a distilled causal-fast variant for
long-horizon chunk-wise rollout and supports camera motion, diverse character
actions, and text-triggered events.~\cite{gao2026lingbotworldv2}

\paragraph{HY-WorldPlay 1.5.}
Tencent Hunyuan's HY-WorldPlay 1.5, also called WorldPlay, is a streaming video
diffusion model for real-time interactive world modeling. Dual Action
Representation encodes user control, Reconstituted Context Memory restores
long-range context, and Context Forcing distills the memory-aware autoregressive
model.~\cite{sun2025worldplay}

\paragraph{Fantasy-World.}
Developed by AMAP at Alibaba Group, Fantasy-World jointly predicts video and 3D
scene structure in a unified feed-forward model. Its integrated reconstruction
and generation blocks couple an appearance branch with an explicit
geometry-consistency branch, using a frozen WanDiT denoiser for
preconditioning.~\cite{dai2025fantasyworld}

\paragraph{InSpatio-World.}
InSpatio Team's InSpatio-World is a real-time 4D world simulator derived from a
reference video. Its spatiotemporal autoregressive architecture combines an
implicit cache for persistent history with explicit spatial constraints that map
interactions to geometrically plausible camera trajectories.~\cite{shen2026inspatioworld}

\paragraph{Lyra 2.}
NVIDIA's Lyra 2.0 produces persistent, explorable 3D worlds by combining
camera-controlled video generation with feed-forward 3D reconstruction. It uses
per-frame geometry to retrieve historical views and establish dense
correspondences, plus self-augmented histories to reduce spatial forgetting and
autoregressive drift.~\cite{shen2026lyra2}

\paragraph{Native-action models.}

\paragraph{SANA-WM.}
NVIDIA's SANA-WM is an open minute-scale controllable world model based on a
hybrid linear diffusion transformer. It combines Gated DeltaNet and softmax
attention, dual-branch 6-DoF camera control, and a two-stage
generation-and-refinement pipeline for efficient long-context synthesis.~\cite{zhu2026sanawm}

\paragraph{DreamX-World.}
Alibaba's AMAP-ML DreamX team presents DreamX-World as a general-purpose
interactive text/image-to-video model. It converts a bidirectional generator
into a few-step autoregressive DiT using causal forcing and distillation, and
adds geometry-guided memory retrieval for revisit consistency and long-horizon
control.~\cite{dreamx2026world}

\paragraph{ABot-World.}
Alibaba's AMAP CV Lab developed ABot-World as an action-conditioned video world
model for real-time, long-horizon closed-loop interaction. It distills a
bidirectional teacher into a causal DiT student and uses LongForcing, raw
keyboard control, and a streaming inference stack to limit autoregressive
drift.~\cite{jiang2026abotworld}

\section{Case Construction Details}
\label{app:implementation_details}

As described in the main paper, each case is built from a sampled initial world
and probe family, then grounded and validated against the generated observation.

\subsection{Construction Process}
\label{app:case_construction_process}

\textbf{Scene Taxonomy Sampling.} We describe each initial world along six
complementary axes. \emph{Environment} defines the global setting through a
hierarchy from a broad domain to a semantic group and then a concrete scene.
\emph{Foreground} identifies the dominant entity or interactable element near
the camera, while \emph{Midground} specifies the connecting or activity region
in which an interaction takes place. \emph{Scene Density}, \emph{Appearance},
and \emph{Perspective} control visible complexity, rendering domain, and viewpoint,
respectively.

\begin{table}[h]
\centering
\scriptsize
\setlength{\tabcolsep}{5pt}
\renewcommand{\arraystretch}{1.28}
\begin{tabularx}{\textwidth}{m{0.16\linewidth}m{0.22\linewidth}m{\dimexpr\linewidth-0.38\linewidth-6\tabcolsep\relax}}
\toprule
\textbf{Axis} & \textbf{Role} & \textbf{Representative structure or values} \\
\midrule
Environment & Global scene setting & Indoor $\rightarrow$ Residential $\rightarrow$ living room, bedroom, \ldots \\
& & Outdoor $\rightarrow$ Natural land $\rightarrow$ forest, mountain trail, \ldots \\
& & Transitional $\rightarrow$ Passage \& threshold $\rightarrow$ tunnel, bridge deck, \ldots \\
\midrule
Foreground & Dominant near-camera entity & Agent; vehicle; articulated object; physical system; dynamic process; \ldots \\
\midrule
Midground & Connecting or activity region & Navigation surface; interaction surface; boundary or portal; transport zone; \ldots \\
\midrule
Scene Density & Visible complexity & Sparse; moderate; dense \\
\midrule
Appearance & Rendering domain & Photorealistic; cinematic realistic; game-engine realistic; stylized 3D; anime \\
\midrule
Perspective & Viewpoint & First-person; third-person \\
\bottomrule
\end{tabularx}
\caption{\textbf{Representative structure of the six-axis scene taxonomy.}
Each case selects one value from every axis. The Environment axis is shown as a
hierarchical path, while representative categories or values summarize the
remaining axes.}
\label{tab:app_scene_taxonomy}
\end{table}
\FloatBarrier

The sampler rejects semantically incompatible axis combinations. A case-specific
layout brief and distinctive visual details then instantiate each accepted tuple
as a spatially coherent, visually distinct world.

\textbf{Probe Family Assignment.} Each initial world is assigned one of the six
probe families defined in the main paper. Assignment follows visible scene
affordances: exploratory cases require a traversable route and stable landmarks;
intentional and physical transitions require an identifiable target and
observable initial state; and persistence cases require a stable layout, a
recognizable return target, or an ongoing process that can evolve while
offscreen.

\textbf{Agentic Case Authoring.} The \textbf{Image Generator} converts the
sampled metadata into a structured prompt and produces the initial observation.
Given this image and the fixed probe family, the \textbf{Image-grounded Planner}
authors a feasible action using the image as the source of truth, without
introducing unseen targets or routes. The \textbf{Case Validator} then checks
target visibility, action feasibility, outcome specificity, and evidence
sufficiency before admitting the case.

\subsection{Case Gallery}
\label{app:case_gallery}
\begin{figure}[p]
    \centering
    \includegraphics[height=0.9\textheight]{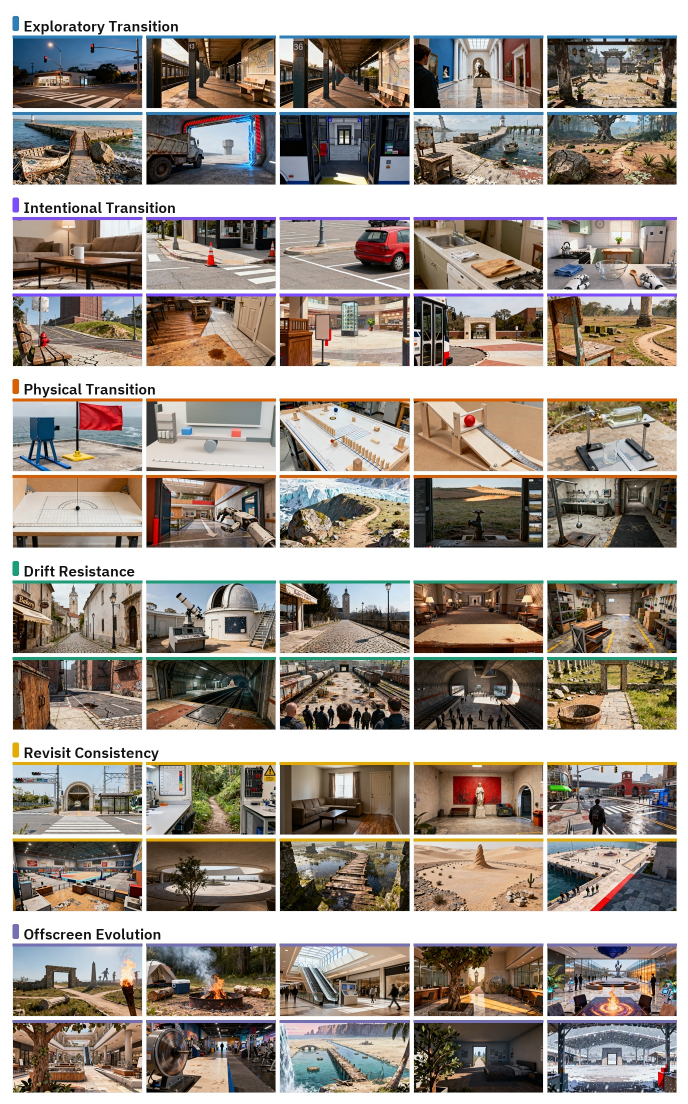}
    \caption{\textbf{Probe-family thumbnail gallery of \ourmodel cases.}
    The gallery shows ten unique sampled initial observations from each probe family.}
    \label{fig:app_case_gallery_by_family}
\end{figure}

Figure~\ref{fig:app_case_gallery_by_family}, shows representative initial observations by
different probe families, where each probe family contains variable cases including different environments, appearances and events.


\FloatBarrier

\section{Evaluation Protocol}
\label{app:evaluation_details}

\subsection{Skill-based Measurement}
\label{app:skill_library}

\paragraph{Evaluation Protocol.}
We evaluate each case through a set of specialized skills, with each skill
providing a normalized score together with the evidence supporting its judgment.
Skill selection depends only on the initial observation, action, and evaluation
setting, and is therefore fixed across models for the same case. Each selected
skill operates on the case-defined action intervals. When mandatory evidence is
missing, the corresponding measurement is marked unavailable rather than
assigned a default score.

For compactness, we use
\[
\operatorname{clip}_{01}(x)=\max(0,\min(1,x)),
\qquad
[x]_+=\max(1,0),
\]
and define the linear ramp as
\[
\operatorname{ramp}(x;\ell,h)=
\begin{cases}
0,&x\leq\ell,\\
1,&x\geq h,\\
(x-\ell)/(h-\ell),&\text{otherwise}.
\end{cases}
\]
Unless a skill specifies a stricter completeness requirement, aggregate scores
use only available numeric components, with their weights renormalized to sum to
one.

The following paragraphs provide detailed descriptions of the skills used in our skill-based evaluation framework.

\paragraph{Render Quality Inspector.}
The Render Quality Inspector evaluates the perceptual and temporal rendering quality of generated images and videos. Following the skill design of VBench~\cite{huang2024vbench}, we assess four complementary aspects: aesthetic quality (AQ), image quality (IQ), human preference (HP), and flickering (FL). Specifically, aesthetic quality is evaluated using a LAION-based aesthetic predictor with CLIP features, image quality is measured by MUSIQ~\cite{ke2021musiq}, and human preference is evaluated using HPSv3. For videos, we sample frames at 2 FPS for aesthetic and image quality evaluation, at 10 FPS for flickering measurement, and uniformly sample at most 20 frames for human preference evaluation.

Let \(X_t\) denote the \(t\)-th decoded RGB frame. Let \(K_{\mathrm{AQ}}\), \(K_{\mathrm{IQ}}\), \(K_{\mathrm{HP}}\), and \(K_{\mathrm{FL}}\) denote the numbers of frames sampled for the corresponding evaluations. The four components are computed as:

$$
\begin{aligned}
\mathrm{AQ}
&=
\frac{1}{K_{\mathrm{AQ}}}
\sum_{t=1}^{K_{\mathrm{AQ}}}
\frac{
\operatorname{LAION}
\bigl(
\operatorname{CLIP}_{\mathrm{L/14}}(X_t)
\bigr)
}{10},
\\
\mathrm{IQ}
&=
\frac{1}{K_{\mathrm{IQ}}}
\sum_{t=1}^{K_{\mathrm{IQ}}}
\frac{
\operatorname{MUSIQ}(X_t)
}{100},
\\
\mathrm{HP}
&=
\operatorname{clip}_{01}
\left(
\frac{1}{K_{\mathrm{HP}}}
\sum_{t=1}^{K_{\mathrm{HP}}}
\frac{
\operatorname{HPSv3}(X_t)
}{10}
\right),
\\
\mathrm{FL}
&=
1-
\frac{1}{255(K_{\mathrm{FL}}-1)}
\sum_{t=1}^{K_{\mathrm{FL}}-1}
\operatorname{MAE}
\left(
X_{t+1},X_t
\right),
\\
\mathrm{RQ}
&=
\frac{1}{4}
\left(
\mathrm{AQ}
+
\mathrm{IQ}
+
\mathrm{HP}
+
\mathrm{FL}
\right).
\end{aligned}
$$

For MUSIQ~\cite{ke2021musiq}, frames whose long edge exceeds 512 pixels are resized while preserving their aspect ratio. If the flicker sampler returns fewer than two frames, we set \(\mathrm{FL}=1\).

\paragraph{Motion Quality Inspector.}
This skill evaluates motion quality through dynamic motion detection and temporal predictability. At 8 FPS, RAFT (or Farneback as a fallback) estimates optical flow between consecutive frames. Let $\mathbf{F}_t=(F_{t,x},F_{t,y})$ denote the flow field between $X_t$ and $X_{t+1}$. We compute the mean magnitude of the top $5\%$ flow values,
\[
M_t
=
\operatorname{mean}
\left(
\operatorname{Top}_{5\%}
\left\{
\sqrt{F_{t,x}^2+F_{t,y}^2}
\right\}
\right),
\qquad
\tau_M
=
6\,\frac{\min(H_f,W_f)}{256},
\]
and define
\[
D
=
\mathbf{1}
\left[
\frac{1}{K-1}
\sum_{t=1}^{K-1}
\mathbf{1}[M_t>\tau_M]
\geq 0.25
\right].
\]
At 10 FPS, AMT-S predicts each intermediate frame from its adjacent frames:
\[
\widehat{X}_{2t+1}
=
\operatorname{AMT\text{-}S}
\left(
X_{2t},X_{2t+2};0.5
\right),
\qquad
S_{\mathrm{tp}}
=
1-
\frac{1}{255K_{\mathrm{int}}}
\sum_{t=1}^{K_{\mathrm{int}}}
\operatorname{MAE}
\left(
\widehat{X}_{2t+1},X_{2t+1}
\right).
\]
The final motion quality score is
\[
\mathrm{MQ}
=
\frac{1}{2}
\left(
D+S_{\mathrm{tp}}
\right).
\]
If fewer than two frames are available, $D=0$; if fewer than three frames are available, $S_{\mathrm{tp}}=0$.

\paragraph{Appearance Consistency Inspector.}
This skill evaluates the temporal consistency of visual appearance in generated videos, capturing both local frame-to-frame discontinuities and accumulated appearance drift. The output video is sampled at 2 FPS, and a normalized CLIP ViT-B/32 embedding is extracted from each sampled RGB frame. Let \(X_t\) denote the \(t\)-th sampled frame and \(K\) denote the total number of sampled frames:
\[
z_t
=
\operatorname{norm}
\left(
\operatorname{CLIP}_{\mathrm{B/32}}(X_t)
\right),
\qquad
t=0,\ldots,K-1.
\]
The appearance consistency score compares each frame with both its immediate predecessor and the first sampled frame:
\[
\mathrm{AC}
=
\frac{1}{2(K-1)}
\sum_{t=1}^{K-1}
\left(
\left[
\cos(z_t,z_{t-1})
\right]_+
+
\left[
\cos(z_t,z_0)
\right]_+
\right),
\]
where \([x]_+=\max(x,0)\). The first term measures local appearance continuity between consecutive frames, while the second measures long-term consistency relative to the initial frame, capturing both abrupt appearance changes and accumulated visual drift. For very short videos, decoding is retried without temporal subsampling to recover additional frames. If fewer than two valid frames remain after decoding, we set \(\mathrm{AC}=1\). The resulting score lies in \([0,1]\), with higher values indicating more consistent visual appearance throughout the video.

\paragraph{Physical Plausibility Inspector.}
This skill evaluates the physical plausibility of generated videos using visual
plausibility and causal fidelity. For visual plausibility, PAVRM, a Qwen3-VL-based
reward model from WBench~\cite{ying2026wbench}, evaluates frames sampled at 2 FPS. Given the logits $\{\lambda_{t,i}\}_{i=1}^{5}$ for frame $X_t$, corresponding to the ordered anchors $\{\mathrm{Perfect},\mathrm{Good},\mathrm{Fair},\mathrm{Poor},\mathrm{Bad}\}$ with weights $w_i\in\{5,4,3,2,1\}$, we first compute a softmax distribution over the five anchors:
\[
p_{t,i}
=
\frac{\exp(\lambda_{t,i})}
{\sum_{j=1}^{5}\exp(\lambda_{t,j})},
\qquad
V_t
=
\frac{1}{5}
\sum_{i=1}^{5}
p_{t,i}w_i.
\]
For $K_V$ valid sampled frames, the video-level visual plausibility score is
\[
V
=
\frac{1}{K_V}
\sum_{t=1}^{K_V}
V_t.
\]

Separately, a VLM evaluates the causal fidelity of the full video sampled at approximately 3 FPS. It assigns an overall causal score $c_0\in\{0,1,2,3\}$, together with dimension-specific scores $c_d\in\{0,1,2,3\}$ for applicable physical phenomena, including fluid dynamics, collision, surface interaction, deformation, wind and weather, reflection, and human physics. Let $\mathcal{D}$ denote the set of applicable dimensions with valid scores. The normalized causal-fidelity score is defined as
\[
C_f
=
\begin{cases}
\displaystyle
\frac{1}{6}
\left(
c_0+
\frac{1}{|\mathcal{D}|}
\sum_{d\in\mathcal{D}}c_d
\right),
& |\mathcal{D}|>0,
\\[12pt]
\displaystyle
\frac{c_0}{3},
& |\mathcal{D}|=0.
\end{cases}
\]
The causal-fidelity evaluation requires at least two valid video frames. When both visual plausibility and causal fidelity are available, the final physical plausibility score is their arithmetic mean:
\[
\mathrm{PP}
=
\frac{V+C_f}{2}.
\]
If fewer than two valid frames are available for causal-fidelity evaluation, $C_f$ is undefined and the final score is set to $\mathrm{PP}=V$. Higher $\mathrm{PP}$ indicates greater physical plausibility.

\paragraph{Viewpoint Trajectory Verifier.}
We follow WBench~\cite{ying2026wbench} for exploratory-transition evaluation. MegaSAM~\cite{li2025megasam} estimates the camera-to-world poses of the generated video, which are then used for trajectory evaluation. After extracting the retained trajectories, let $e_t$ and $e_r$ denote the mean translation and angular errors, and let $L$ and $\Theta$ denote the predicted translation path length and accumulated rotation in degrees. The navigation accuracy is
\[
\widehat e_t
=
\min\!\left(
\frac{e_t}{\max(L,0.5)},1
\right),
\qquad
\widehat e_r
=
\min\!\left(
\frac{e_r}{\max(\Theta,10)},1
\right),
\qquad
\mathrm{Acc}
=
1-\frac{1}{2}
\left(
\widehat e_t+\widehat e_r
\right).
\]

For consistency, we compare trajectories from symmetric action pairs
$W\!\leftrightarrow\!S$, $A\!\leftrightarrow\!D$,
left$\!\leftrightarrow\!$right, and up$\!\leftrightarrow\!$down. Each pair is aligned to a common initial pose, with opposing actions mirrored before comparison. For comparable pair $k$, let $e_{t,k}^{c}$ and $e_{r,k}^{c}$ denote the mean translation and angular errors, and $\bar L_k$ and $\bar\Theta_k$ the corresponding mean path length and rotation. We compute
\[
\widehat e_{t,k}^{c}
=
\min\!\left(
\frac{e_{t,k}^{c}}{\max(\bar L_k,0.5)},1
\right),
\qquad
\widehat e_{r,k}^{c}
=
\min\!\left(
\frac{e_{r,k}^{c}}{\max(\bar\Theta_k,10)},1
\right),
\]
and
\[
\mathrm{Con}
=
1-
\frac{1}{2|\mathcal P|}
\sum_{k\in\mathcal P}
\left(
\widehat e_{t,k}^{c}
+
\widehat e_{r,k}^{c}
\right),
\qquad
\mathrm{Nav}
=
\frac{1}{2}
\left(
\mathrm{Acc}+\mathrm{Con}
\right),
\]
where $\mathcal P$ is the set of comparable action pairs. When no comparable pair exists, we set $\mathrm{Con}=1$. At least two valid poses are required to report a navigation score.
\paragraph{Intentional Change Verifier.}
This skill evaluates whether an intentional change specified by the action is
correctly realized in the generated video. It consists of an Analyze Agent and a
Verify Agent. The Analyze Agent identifies the target entity, relation, or event
and its expected transition from the action specification. The Verify Agent then
compares the generated video with this expected outcome and produces eight
normalized judgments $Q^{\operatorname{IC}}_1,\ldots,Q^{\operatorname{IC}}_8
\in\{0, 0.25, 0.5, 0.75, 1\}$, covering target visibility, visible transition, intended change,
target specificity, final-state correctness, anchor preservation, absence of
extra events or resets, and overall judgeability. The final score is
\[
\begin{aligned}
E_{\mathrm{chg}}
&=
0.40Q^{\operatorname{IC}}_2
+
0.60Q^{\operatorname{IC}}_3,
\\
P_{\mathrm{int}}
&=
0.60Q^{\operatorname{IC}}_6
+
0.40Q^{\operatorname{IC}}_7,
\\
C_{\mathrm{int}}
&=
0.30E_{\mathrm{chg}}
+
0.25Q^{\operatorname{IC}}_5
+
0.20Q^{\operatorname{IC}}_4
+
0.25P_{\mathrm{int}},
\\
\mathrm{IC}
&=
Q^{\operatorname{IC}}_1
Q^{\operatorname{IC}}_8
C_{\mathrm{int}}.
\end{aligned}
\]
This design ensures that a strong change receives a high score only when the
target is visible and the resulting transition is clearly judgeable.

\paragraph{Physical Response Verifier.}
This skill evaluates whether an intervention produces the expected physical
response. It consists of an Analyze Agent and a Verify Agent. The Analyze Agent
identifies the physical setup, intervention, and expected response from the
action specification. The Verify Agent then compares the generated video with
this expected outcome and produces eight normalized judgments
$Q^{\operatorname{PR}}_1,\ldots,Q^{\operatorname{PR}}_8\in\{0, 0.25, 0.5, 0.75, 1\}$, covering
setup visibility, intervention visibility, response presence, condition
following, causal order, anchor preservation, absence of physical artifacts,
and judgeability. The final score is
\[
\begin{aligned}
G_{\mathrm{intv}}
&=
\frac{1}{2}
\left(
Q^{\operatorname{PR}}_1+
Q^{\operatorname{PR}}_2
\right),
\\
E_{\mathrm{resp}}
&=
0.45Q^{\operatorname{PR}}_3+
0.35Q^{\operatorname{PR}}_4+
0.20Q^{\operatorname{PR}}_5,
\\
P_{\mathrm{phys}}
&=
0.55Q^{\operatorname{PR}}_6+
0.45Q^{\operatorname{PR}}_7,
\\
C_{\mathrm{phys}}
&=
0.55E_{\mathrm{resp}}+
0.25P_{\mathrm{phys}}+
0.20Q^{\operatorname{PR}}_5,
\\
\mathrm{PR}
&=
G_{\mathrm{intv}}
Q^{\operatorname{PR}}_8
C_{\mathrm{phys}}.
\end{aligned}
\]
This design ensures that the physical response is rewarded only when the
intervention is visible, the expected response occurs in the correct causal
order, and the outcome is clearly judgeable.

\paragraph{Physical Law Validator.}
Beyond evaluating general physical plausibility, this skill verifies whether specific physical transitions follow their expected physical laws. For each supported transition, a dedicated physical-law engine is selected to perform case-specific diagnostics, including projectile motion, periodic and bouncing motion, event ordering, vertical motion, deceleration, reversal, settling, and collision energy loss. The video is sampled at 2 FPS and analyzed using optical flow and motion trajectories. For each registered physical-law check $\ell$, the corresponding engine produces a normalized score $L_\ell\in[0,1]$. Only checks applicable to the selected engine are evaluated, and the overall physical-law score is
\[
\mathrm{PL}
=
\frac{1}{|\mathcal{L}|}
\sum_{\ell\in\mathcal{L}} L_\ell,
\]
where $\mathcal{L}$ denotes the set of valid physical-law checks. At least three valid frames are required for physical-law evaluation. The optical-flow-based diagnostics assume a relatively static camera, static background, moderate motion, and a dominant textured subject, and are therefore used as auxiliary physical-law verification.

\paragraph{Drift Degradation Analyzer.}
This skill evaluates quality degradation over long rollouts. The rollout is
partitioned into case-defined action chunks $C_0,\ldots,C_{N-1}$, with equal-duration
boundaries used as a fallback. Each chunk must have valid render, motion,
appearance, and physical-plausibility scores:
\[
q_j=
\frac{1}{4}
\left(
\mathrm{RQ}_j+\mathrm{MQ}_j+\mathrm{AC}_j+\mathrm{PP}_j
\right).
\]
Let $x_j=j/(N-1)$ and $\beta$ be the least-squares slope of $q_j$ with respect to
$x_j$. We define trend, retention, worst-case quality, and stability as
\[
\begin{aligned}
T&=\operatorname{clip}_{01}
\left(1-\operatorname{ramp}(-\beta;0.02,0.15)\right),\\
R&=\operatorname{clip}_{01}
\left(\frac{q_{N-1}}{\max(q_0,10^{-6})}\right),\\
W&=\min_{0\leq j<N}q_j,\\
S&=\operatorname{clip}_{01}
\left(1-2\,\operatorname{std}_{\mathrm{pop}}(q_0,\ldots,q_{N-1})\right).
\end{aligned}
\]
For semantic drift, the normalized mean CLIP B/32 embedding of each chunk is
denoted by $z_j$. We compute
\[
c_j=
\frac{1}{2}
\left(
[\cos(z_j,z_{j-1})]_+
+
[\cos(z_j,z_0)]_+
\right),
\qquad
C=
\frac{1}{2}
\left(
\frac{1}{N-1}\sum_{j=1}^{N-1}c_j
+
\min_{1\leq j<N}c_j
\right).
\]
The final score is
\[
\mathrm{Drift}
=
0.35C+0.20T+0.15R+0.15W+0.15S.
\]
If any required chunk score is unavailable, the drift score is unavailable.
\paragraph{Return Consistency Verifier.}
We design our return consistency verifier using return action pairs. Specifically, we uniformly sample at most 12 temporally aligned RGB frames from the full video as a chunk, and we construct paired reference and return chunks for paired actions. For example, two chunks for forward action and backward action, respectively. Considering the central-paired frames $X_a$ and $X_b$ from paired chunks, we calculate the paired similarity score $s(a,b)$ as following:
\[
\begin{aligned}
G(a,b)&=\tfrac13\left[
\operatorname{clip}_{01}\!\left(\tfrac{\operatorname{NCC}(X_a,X_b)+1}{2}\right)
+\operatorname{clip}_{01}\!\left(
\tfrac{\operatorname{corr}(h_a,h_b)+1}{2}\right)
+\operatorname{IoU}(\operatorname{Canny}(X_a),\operatorname{Canny}(X_b))\right],\\
P(a,b)&=1-\operatorname{ramp}\!\left(
\tfrac{\operatorname{MAE}(X_a,X_b)}{255};0.05,0.35\right),\\
E(a,b)&=[\cos(z_a,z_b)]_+,\\
s(a,b)&=0.25G(a,b)+0.25P(a,b)+0.50E(a,b),
\end{aligned}
\]
where $h_a,h_b$ are normalized 32-bin histograms and $z_A,z_B$ are normalized
CLIP B/32 embeddings. Let $\mathcal S_k$ contain all cross-interval
frame-level similarities for each chunk pair $k$, we define pair and multi-pair consistency as:
\[
\begin{aligned}
p_k &= \tfrac{1}{2}\left(\max \mathcal{S}_k
      + \operatorname{mean}\mathcal{S}_k\right),\\
L&=\tfrac12\left(\operatorname{mean}_kp_k+\min_kp_k\right).
\end{aligned}
\]
We then define the return consistency score from $L$.

\paragraph{Offscreen Evolution Verifier.}
This skill evaluates whether an object or process continues to evolve plausibly
while temporarily outside the visible scene. It consists of an Analyze Agent and
a Verify Agent. The Analyze Agent identifies the target process, its visible
states, and the expected offscreen evolution. The Verify Agent compares the
generated video with this expected evolution and produces eight normalized
judgments $Q^{\operatorname{OE}}_1,\ldots,Q^{\operatorname{OE}}_8\in\{0, 0.25, 0.5, 0.75, 1\}$,
covering pre-visibility, offscreen absence, return visibility, anchor
preservation, absence of reset, visible state change, plausible evolution, and
judgeability. The final score is
\[
\begin{aligned}
V_{\mathrm{off}}
&=
\frac{1}{3}
\left(
Q^{\operatorname{OE}}_1+
Q^{\operatorname{OE}}_2+
Q^{\operatorname{OE}}_3
\right),
\\
G_{\mathrm{off}}
&=
\frac{1}{2}
\left(
Q^{\operatorname{OE}}_5+
Q^{\operatorname{OE}}_8
\right),
\\
E_{\mathrm{off}}
&=
0.40Q^{\operatorname{OE}}_6+
0.60Q^{\operatorname{OE}}_7,
\\
C_{\mathrm{off}}
&=
0.35Q^{\operatorname{OE}}_4+
0.65E_{\mathrm{off}},
\\
\mathrm{OE}
&=
V_{\mathrm{off}}
G_{\mathrm{off}}
C_{\mathrm{off}}.
\end{aligned}
\]
This design requires the process to remain consistent across the offscreen
interval and to return with a plausible state evolution rather than a reset.

\paragraph{Model-level aggregation.}
For each probe family $c$, its capability score is computed as the mean score
over all cases belonging to that family. Let $\mathcal{C}_c$ denote the set of
applicable cases in family $c$, and let $S_k$ denote the final score of case $k$.
Then
\[
K_c
=
\frac{1}{|\mathcal{C}_c|}
\sum_{k\in\mathcal{C}_c} S_k.
\]
The overall model score is computed directly over all valid cases:
\[
K_{\mathrm{overall}}
=
\frac{1}{|\mathcal{C}|}
\sum_{k\in\mathcal{C}} S_k,
\]
where $\mathcal{C}$ denotes the set of all valid cases. Cases with unavailable
scores are excluded from the corresponding aggregation.

\subsection{Skill Reasoning Examples}
\label{app:case_examples}

\begin{figure*}[p]
    \centering

    \begin{subfigure}{\linewidth}
        \centering
        \includegraphics[width=\linewidth]{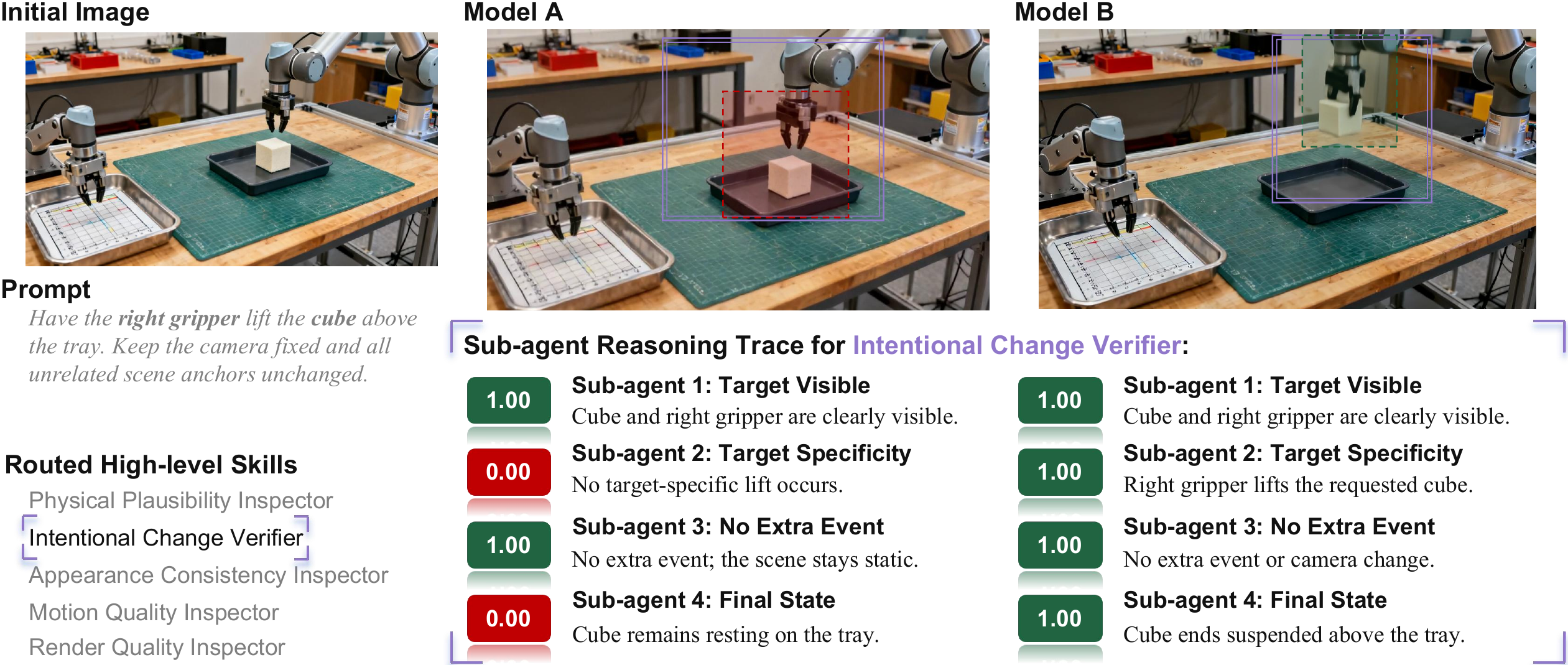}
        \label{fig:case1}
    \end{subfigure}

    \vspace{0.5em}

    \begin{subfigure}{\linewidth}
        \centering
        \includegraphics[width=\linewidth]{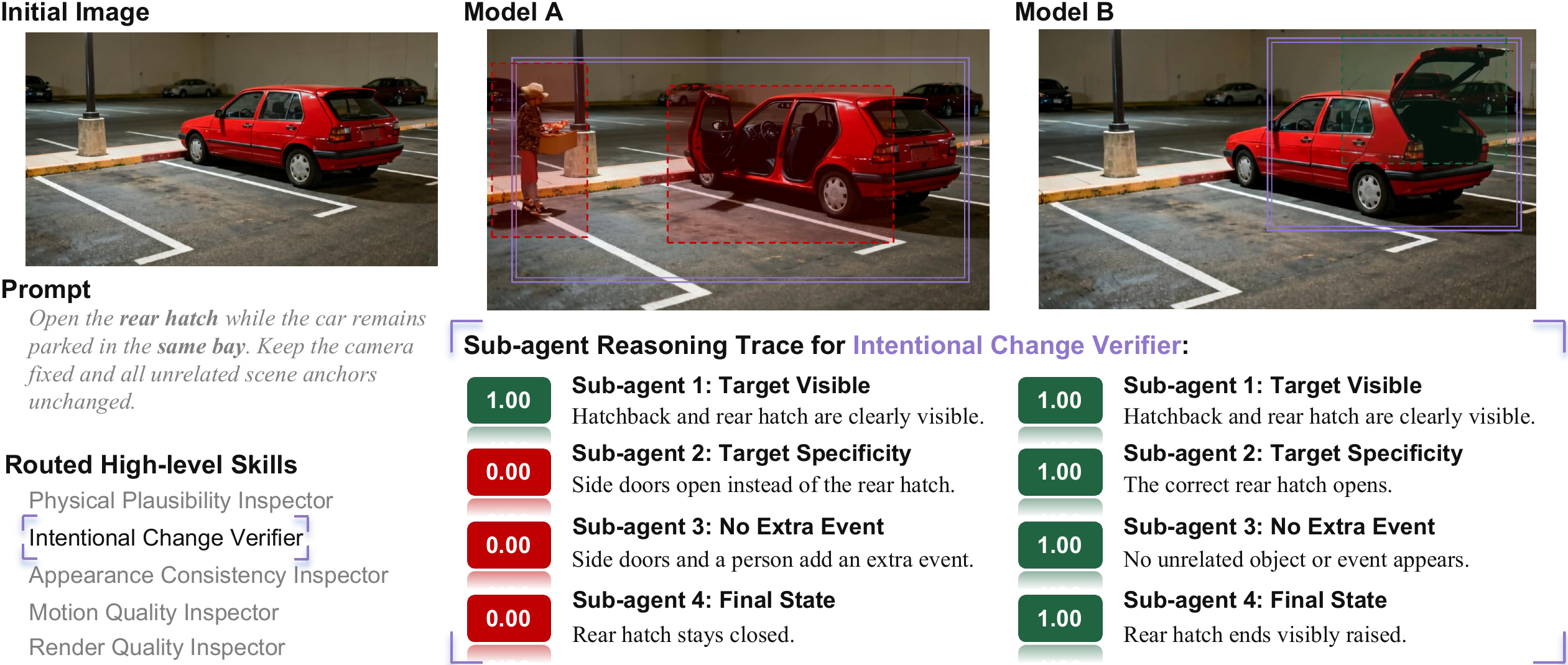}
        \label{fig:case2}
    \end{subfigure}

    \vspace{0.5em}

    \begin{subfigure}{\linewidth}
        \centering
        \includegraphics[width=\linewidth]{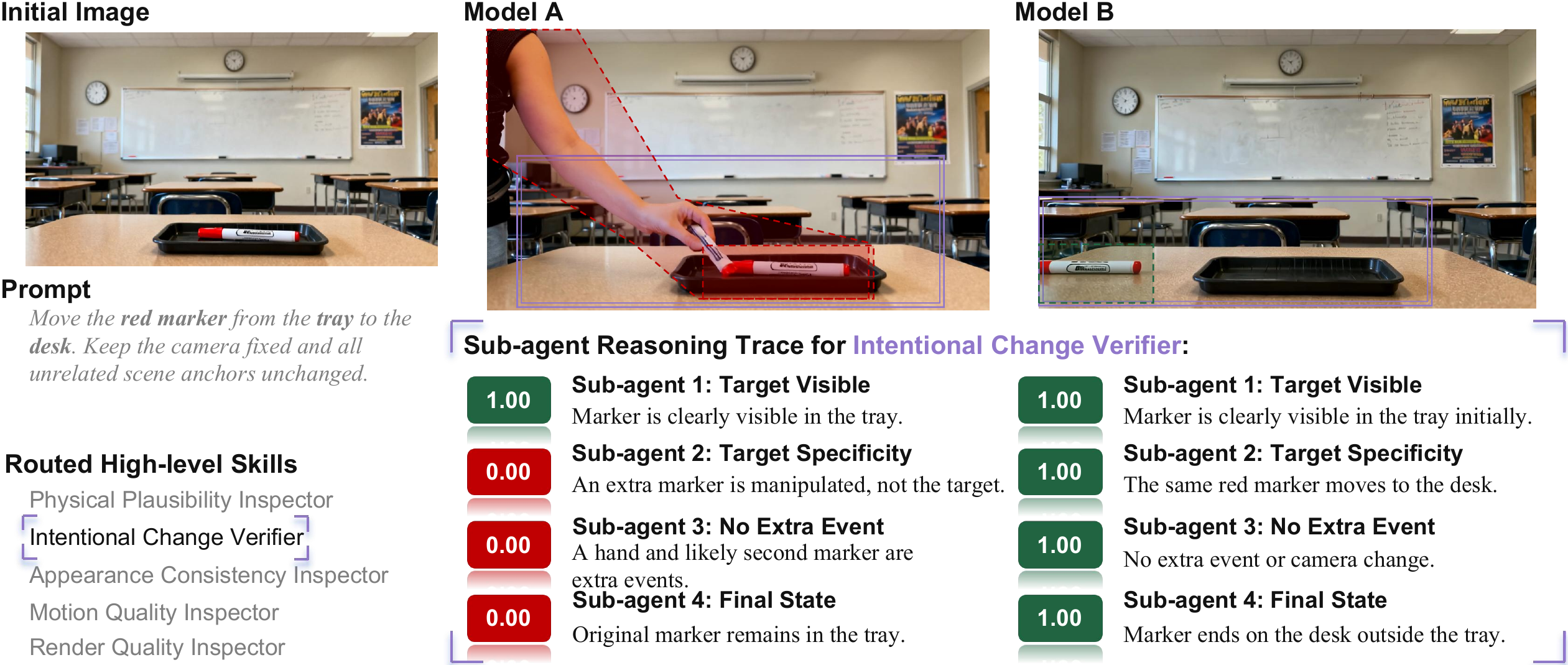}
        \label{fig:case3}
    \end{subfigure}

    \caption{
        \textbf{Representative reasoning traces produced by \ourmodel. (A)}
    }
    \label{fig:cases1-3}
\end{figure*}

\begin{figure*}[p]
    \centering

    \begin{subfigure}{\linewidth}
        \centering
        \includegraphics[width=\linewidth]{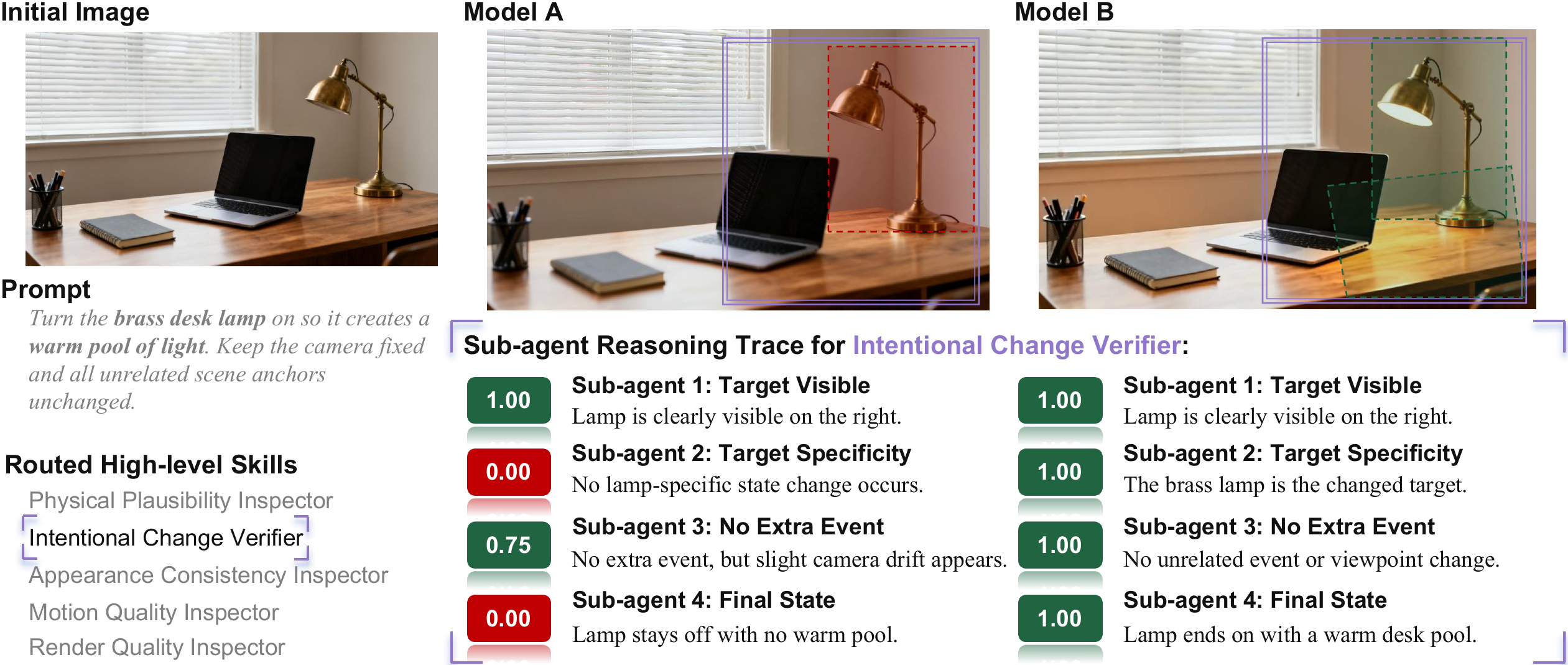}
        \label{fig:case4}
    \end{subfigure}

    \vspace{0.5em}

    \begin{subfigure}{\linewidth}
        \centering
        \includegraphics[width=\linewidth]{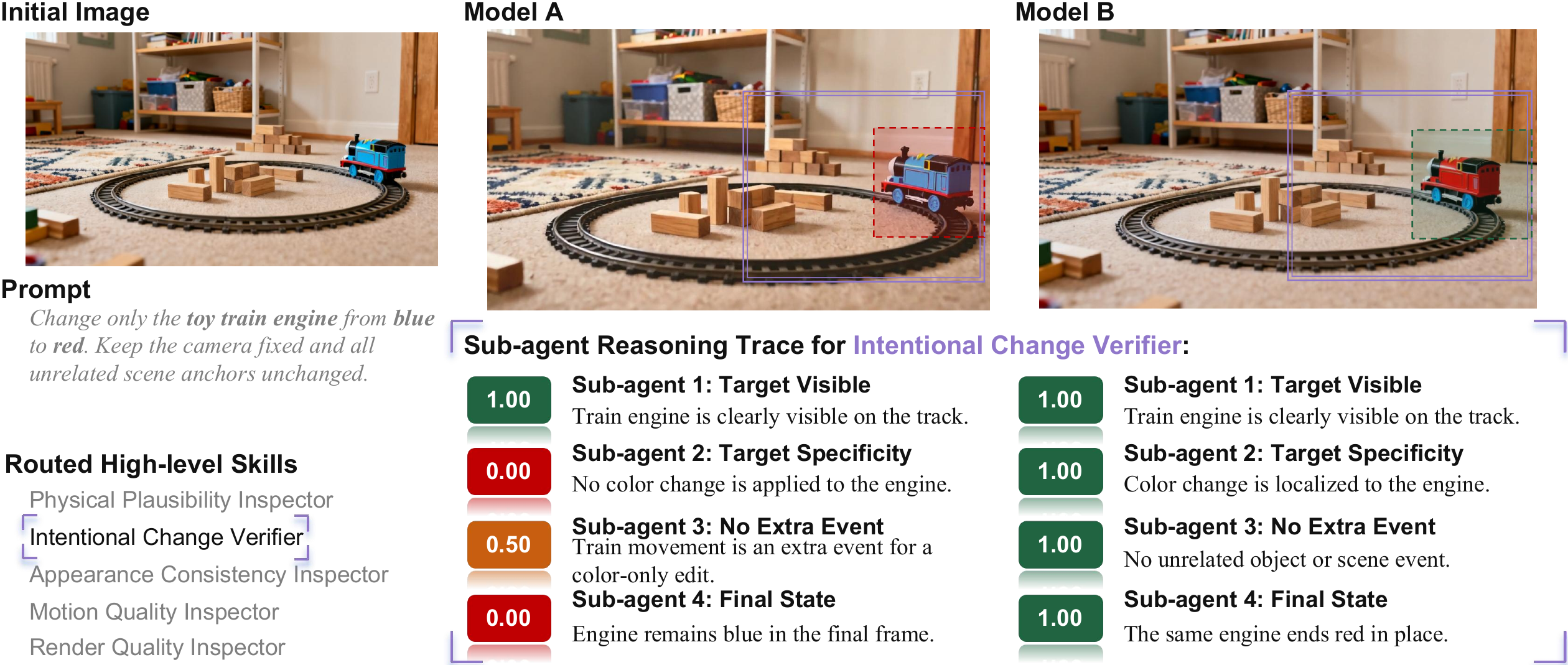}
        \label{fig:case5}
    \end{subfigure}

    \vspace{0.5em}

    \begin{subfigure}{\linewidth}
        \centering
        \includegraphics[width=\linewidth]{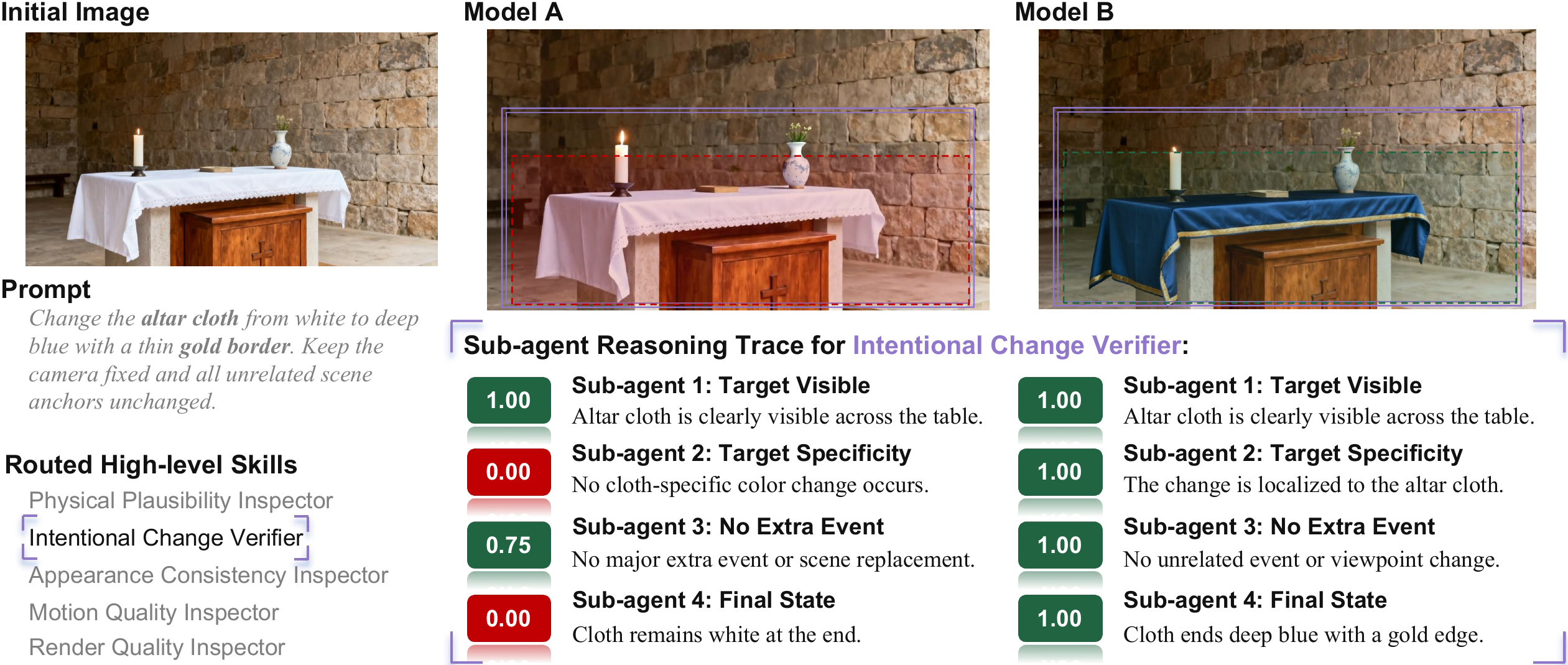}
        \label{fig:case6}
    \end{subfigure}

    \caption{
        \textbf{Representative reasoning traces produced by \ourmodel. (B)}
    }
    \label{fig:cases4-6}
\end{figure*}

\begin{figure*}[p]
    \centering

    \begin{subfigure}{\linewidth}
        \centering
        \includegraphics[width=\linewidth]{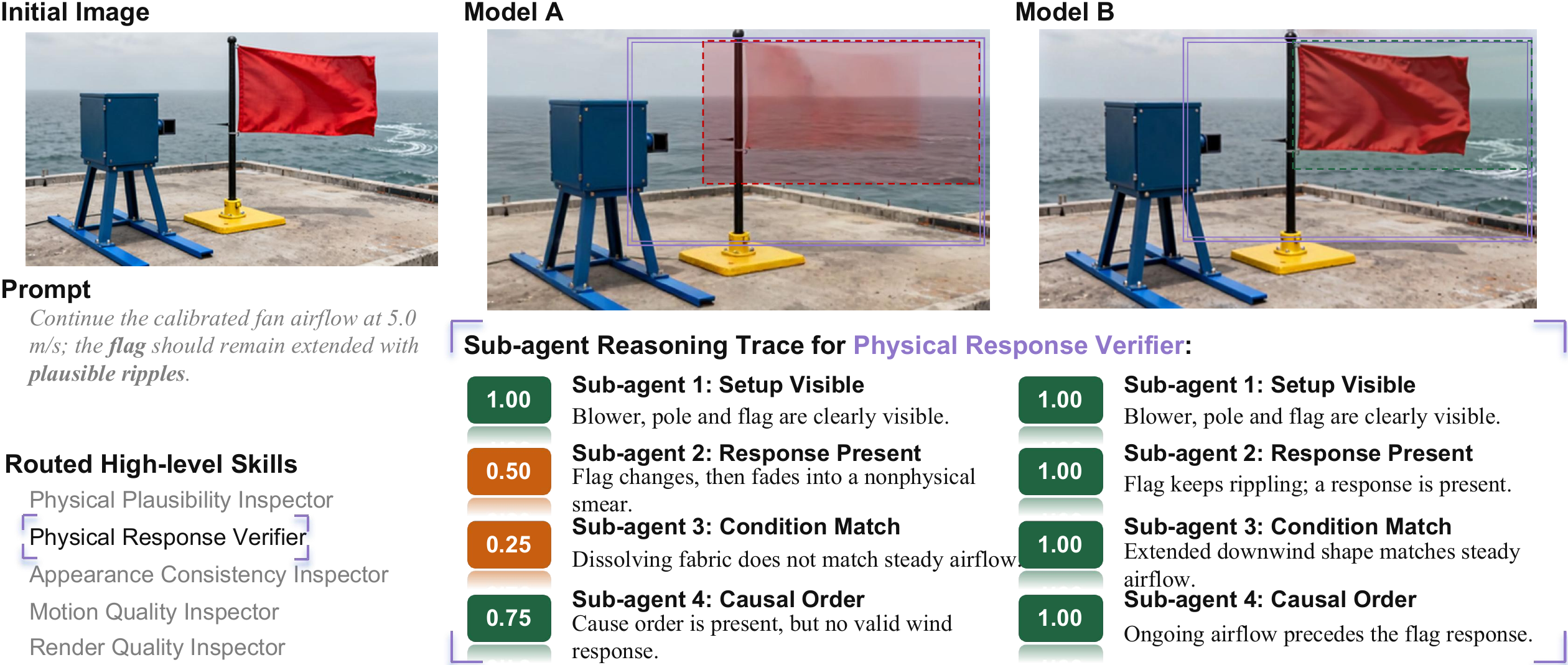}
        \label{fig:case7}
    \end{subfigure}

    \vspace{0.5em}

    \begin{subfigure}{\linewidth}
        \centering
        \includegraphics[width=\linewidth]{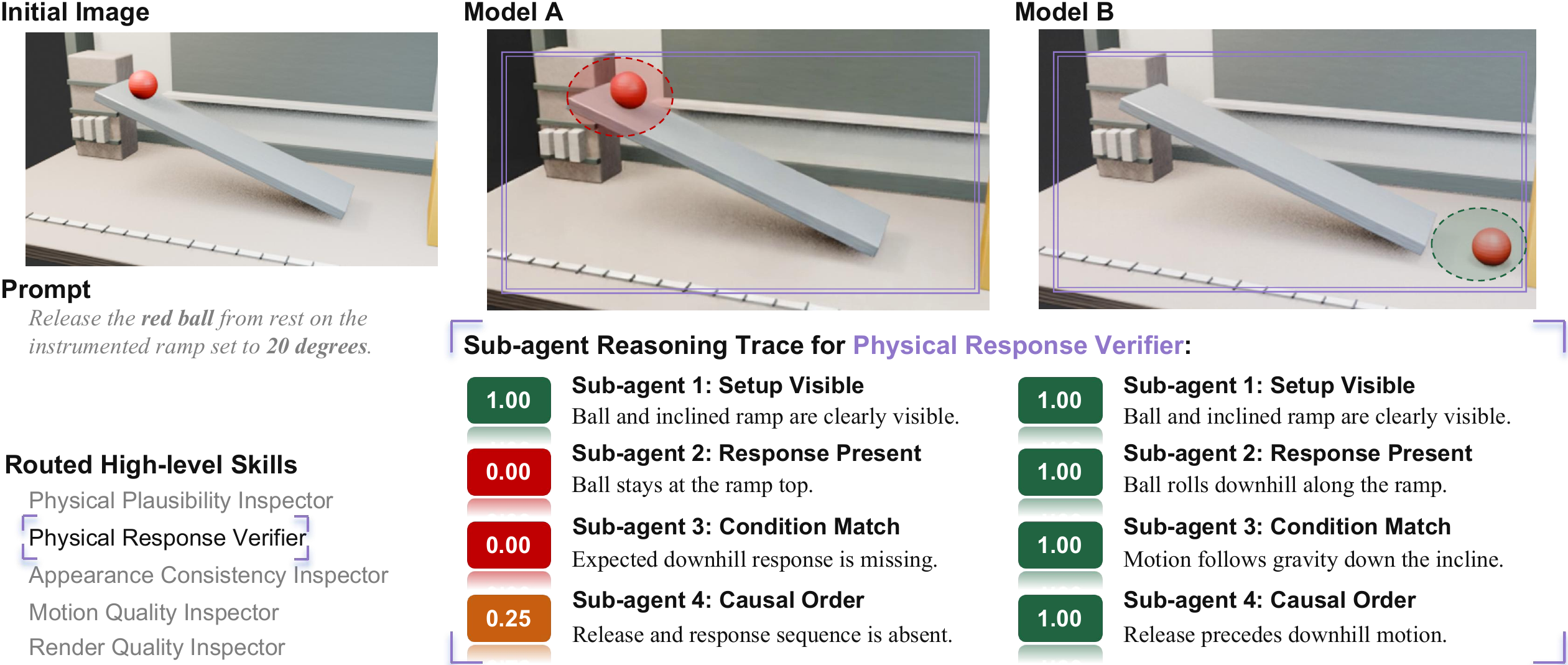}
        \label{fig:case8}
    \end{subfigure}

    \vspace{0.5em}

    \begin{subfigure}{\linewidth}
        \centering
        \includegraphics[width=\linewidth]{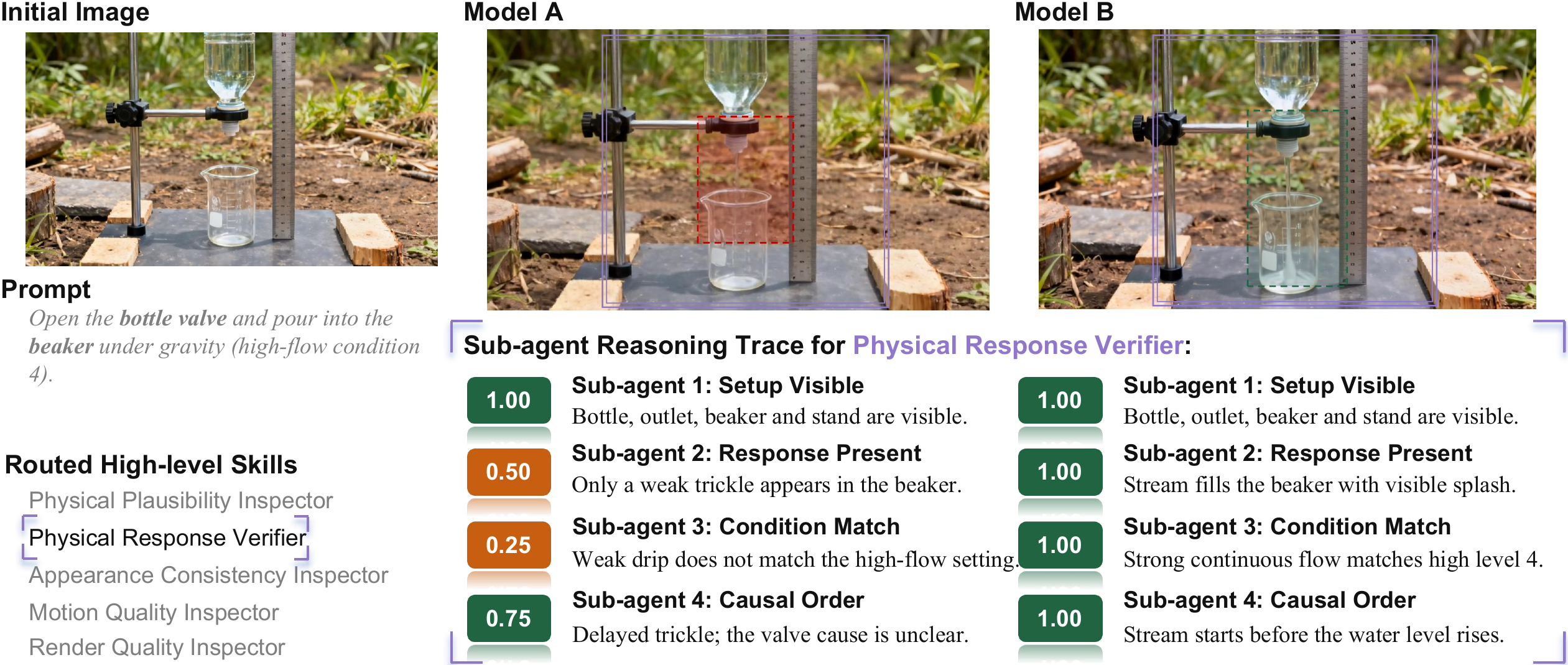}
        \label{fig:case9}
    \end{subfigure}

    \caption{
        \textbf{Representative reasoning traces produced by \ourmodel. (C)}
    }
    \label{fig:cases7-9}
\end{figure*}

We provide representative cases to illustrate how the proposed evaluation
framework analyzes generated videos. Each case shows the initial observation,
the action specification, the routed high-level skills, and the corresponding
reasoning traces for different models. The reasoning traces further expose the
decomposed sub-questions and their supporting evidence, making the evaluation
process and the resulting judgments transparent. Figures~\ref{fig:cases1-3}
--\ref{fig:cases7-9} present representative examples covering intentional
change and physical response evaluation.

\FloatBarrier